%% file: main.tex
\def\year{2022}\relax
\documentclass[letterpaper]{article} %
\usepackage{aaai22}  %
\usepackage{times}  %
\usepackage{helvet}  %
\usepackage{courier}  %
\usepackage[hyphens]{url}  %
\usepackage{graphicx} %
\usepackage{natbib}  %
\usepackage{caption} %
\DeclareCaptionStyle{ruled}{labelfont=normalfont,labelsep=colon,strut=off} %
\usepackage{algorithm}
\usepackage{algorithmic}

\usepackage{newfloat}
\usepackage{listings}
\floatstyle{ruled}
\newfloat{listing}{tb}{lst}{}
\floatname{listing}{Listing}

\usepackage{amsfonts}       %
\usepackage{nicefrac}       %
\usepackage{microtype}      %
\usepackage{xcolor}         %
\usepackage{amsmath}

\usepackage{wrapfig}
\usepackage{graphicx}
\usepackage{grffile}
\usepackage{subcaption}
\usepackage{numprint}
\npthousandsep{,}

\usepackage[nameinlink,capitalize]{cleveref}
\crefname{section}{\S}{\S}
\Crefname{section}{\S}{\S}
\crefname{appendix}{App.}{Apps.}
\Crefname{appendix}{App.}{Apps.}
\crefname{theorem}{Thm.}{Thms.}
\Crefname{theorem}{Thm.}{Thms.}
\crefname{proposition}{Prop.}{Props.}
\Crefname{proposition}{Prop.}{Props.}

\newtheorem{proposition}{Proposition}

\newcommand{\name}{RareGAN}

\newcommand{\mnist}{MNIST}
\newcommand{\cifar}{CIFAR10}

\newcommand{\djs}{d_{\textrm{JS}}}
\newcommand{\dw}{d_{\textrm{W}}}
\newcommand{\dwone}{d_{\textrm{W1}}}

\newcommand{\lgan}{\mathcal{L}_{\textrm{GAN}}}
\newcommand{\lganjs}{\mathcal{L}^{\textrm{JS}}_{\textrm{GAN}}}
\newcommand{\lganw}{\mathcal{L}^{\textrm{W}}_{\textrm{GAN}}}
\newcommand{\lganjsweighted}{\mathcal{L}^{\textrm{JS}'}_{\textrm{GAN}}}
\newcommand{\lganwweighted}{\mathcal{L}^{\textrm{W}'}_{\textrm{GAN}}}
\newcommand{\lclassification}{\mathcal{L}_{\textrm{classification}}}

\newcommand{\Eb}{\mathbb{E}}

\newcommand{\bra}[1]{\left( #1 \right)}
\newcommand{\brb}[1]{\left[ #1 \right]}
\newcommand{\brc}[1]{\left\{ #1 \right\}}

\newcommand{\brl}[1]{\left\lVert #1 \right\rVert_L}

\newcommand{\dataset}{\mathcal{D}}
\newcommand{\numsample}{{n}}
\newcommand{\weight}{{w}}
\newcommand{\support}[1]{{\textrm{Support}\bra{#1}}}
\newcommand{\p}{{p}}
\newcommand{\plabel}{{p_{{l}}}}
\newcommand{\psamplelabel}{p_{xl}}
\newcommand{\pgeneratesamplelabel}{\hat{p}_{xl}}
\newcommand{\pgenerate}{{\hat{p}}}

\newcommand{\q}{{q}}
\newcommand{\qgenerate}{{\hat{q}}}

\newcommand{\rarefrc}{\alpha}
\newcommand{\rarefrclearned}{\hat{\alpha}}
\newcommand{\prare}{{p_{\mathrm{r}}}}
\newcommand{\pcommon}{{p_{\mathrm{c}}}}
\newcommand{\pgeneraterare}{{\hat{p}_{\mathrm{r}}}}
\newcommand{\pgeneratecommon}{{\hat{p}_{\mathrm{c}}}}
\newcommand{\pgenerateraresupport}{{\hat{p}'_{\mathrm{r}}}}
\newcommand{\pgeneratecommonsupport}{{\hat{p}'_{\mathrm{c}}}}
\newcommand{\requestbgt}{B}
\newcommand{\numstage}{S}

\newcommand{\hrare}{{h_{\mathrm{r}}}}
\newcommand{\hgeneraterare}{{\hat{h}_{\mathrm{r}}}}

\newcommand{\myparatightest}[1]{\noindent\textbf{{#1.}}~}

\usepackage{mathtools}
\usepackage{booktabs}

\newcounter{packednmbr}
\newenvironment{packedenumerate}{\begin{list}{\thepackednmbr.}{\usecounter{packednmbr}\setlength{\itemsep}{0.5pt}\addtolength{\labelwidth}{10pt}\setlength{\leftmargin}{\labelwidth}\setlength{\listparindent}{\parindent}\setlength{\parsep}{0pt}\setlength{\topsep}{0pt}}}{\end{list}}

\title{\name{}: Generating Samples for Rare Classes}
\author{Zinan Lin, Hao Liang, Giulia Fanti, Vyas Sekar}
\affiliations{Carnegie Mellon University\\
zinanl@andrew.cmu.edu, hl106@rice.edu, gfanti@andrew.cmu.edu, vsekar@andrew.cmu.edu}

\usepackage{bibentry}

\begin{document}

\maketitle

\input{tex/abstract}
\input{tex/motivation}

\input{tex/background}
\input{tex/approach}

\input{tex/experiments}

\input{tex/discussion}

\input{tex/ack}

\bibliography{aaai22}

\clearpage
\onecolumn
\section*{Appendix}
\appendix

\input{tex/app_proof}
\clearpage
\input{tex/experiment_details}
\clearpage
\input{tex/app_bgt_frc}

\end{document}

%% file: tex/abstract.tex
\begin{abstract}
We study the problem of learning generative adversarial networks (GANs) for a rare class of an unlabeled dataset subject to a labeling budget.  This problem is motivated from practical applications in domains including  security	(e.g., synthesizing packets for DNS  amplification attacks), systems and networking (e.g., synthesizing workloads that trigger high resource usage), and machine learning   (e.g., generating images from a rare class).  
Existing approaches are unsuitable, either requiring fully-labeled datasets or sacrificing the fidelity of the rare class for that of the common classes.  
We propose \name{}, a novel  synthesis of three key ideas: (1) extending conditional GANs to  use labelled \emph{and} unlabelled data for better generalization; (2)  an active learning approach that requests the most useful labels; and (3) a weighted loss function to favor learning the rare class. We show  that \name{} achieves a better fidelity-diversity tradeoff on the rare class than prior work  across different applications, budgets, rare class fractions, GAN losses, and architectures\footnote{This work builds on our previous workshop paper (Section 4.2 of \cite{lin2019towards}).}. 
\end{abstract}

%% file: tex/motivation.tex
\section{Introduction}
\label{sec:intro}

Many practitioners in diverse domains such as security, networking, and systems require samples from \emph{rare} classes. For example, operators often want to generate %
queries that force servers to send undesirable responses
\cite{moon2021accurately}, or generate packets that trigger high CPU/memory usage  or processing delays for performance evaluation \cite{petsios2017slowfuzz}. 

Prior domain-specific solutions to these problems rely heavily on prior knowledge (e.g., source code) of the systems, which is often unavailable \cite{lin2019towards}. Indeed, in response to a recent executive order \emph{Improving the Nation's Cybersecurity}\footnote{\url{https://www.federalregister.gov/documents/2021/05/17/2021-10460/improving-the-nations-cybersecurity}}, the U.S. National Institute of Standards and Technology published guidance highlighting the importance of creating `black box' tests for device and software security that do not rely on the implementation or source code of systems \cite{black2021guidelines}.

Given the success of generative adversarial networks (GANs) \cite{goodfellow2014generative} on data generation, we ask if we can use GANs to generate samples from a rare class (e.g., attack packets, packets that trigger high CPU usage) without requiring prior knowledge about the systems. Note that there are two unique characteristics in our problem:
\begin{packedenumerate}
	\item [C1.] \emph{High labeling cost.} Labels (whether a sample belongs to the rare class) are often not available a priori, and getting labels is often resource intensive. For example, for a new system, we often do not know a priori which packets will trigger high CPU usage, and evaluating the CPU usage of a packet (for labeling it) can be time consuming.
	\item [C2.] \emph{Rare class only.} We only need samples from the \emph{rare} class (e.g., attack packets); system operators are often less concerned about \emph{common} class samples (e.g., benign packets).
\end{packedenumerate}
\noindent To the best of our knowledge, no prior GAN paper considers both constraints. Prior related work (see \cref{sec:related}) often assumes that the labels are available (failing C1), or tries to generate both rare \emph{and} common samples, which sacrifices the fidelity on the rare class (failing C2). We will see in \cref{sec:approach} that these new characteristics bring unique challenges. 

\myparatightest{Contributions}
We propose \name{}, a generative model for rare data classes, given an unlabeled dataset and a labeling budget.  
It combines three %
ideas:
(1) It %
modifies existing conditional GANs \cite{odena2017conditional}
to use both labelled \emph{and} unlabelled data for better generalization. %
(2) It uses active learning to label samples; we show theoretically that unlike prior work \cite{xie2019learning}, our implementation does not bias the learned rare class distribution.
(3) It uses a weighted loss function that favors learning the rare class over the common class; we propose efficient optimization techniques for realizing this reweighting.

\begin{table}[t]
	\centering
	\begin{tabular}{c|ccc}
		\toprule
		& Budget$\downarrow$ & Fidelity$\downarrow$& Diversity$\uparrow$ \\
		\midrule
		AmpMAP & \numprint{14788089} & 16.60 & $1.68\%$\\
		\name{} (ours) & \textbf{\numprint{200000}} & \textbf{4.16} & $\boldsymbol{98.07\%}$\\
		\bottomrule
	\end{tabular}
	\caption{RareGAN achieves better fidelity and diversity with lower budget on DNS amplification attacks than domain-specific techniques. See \cref{sec:setup} for the definition of metrics.}
	\label{tbl:ampmap}
	\vspace{-0.2cm}
\end{table}

\begin{figure}[t]
	\centering
	\includegraphics[width=1.\linewidth]{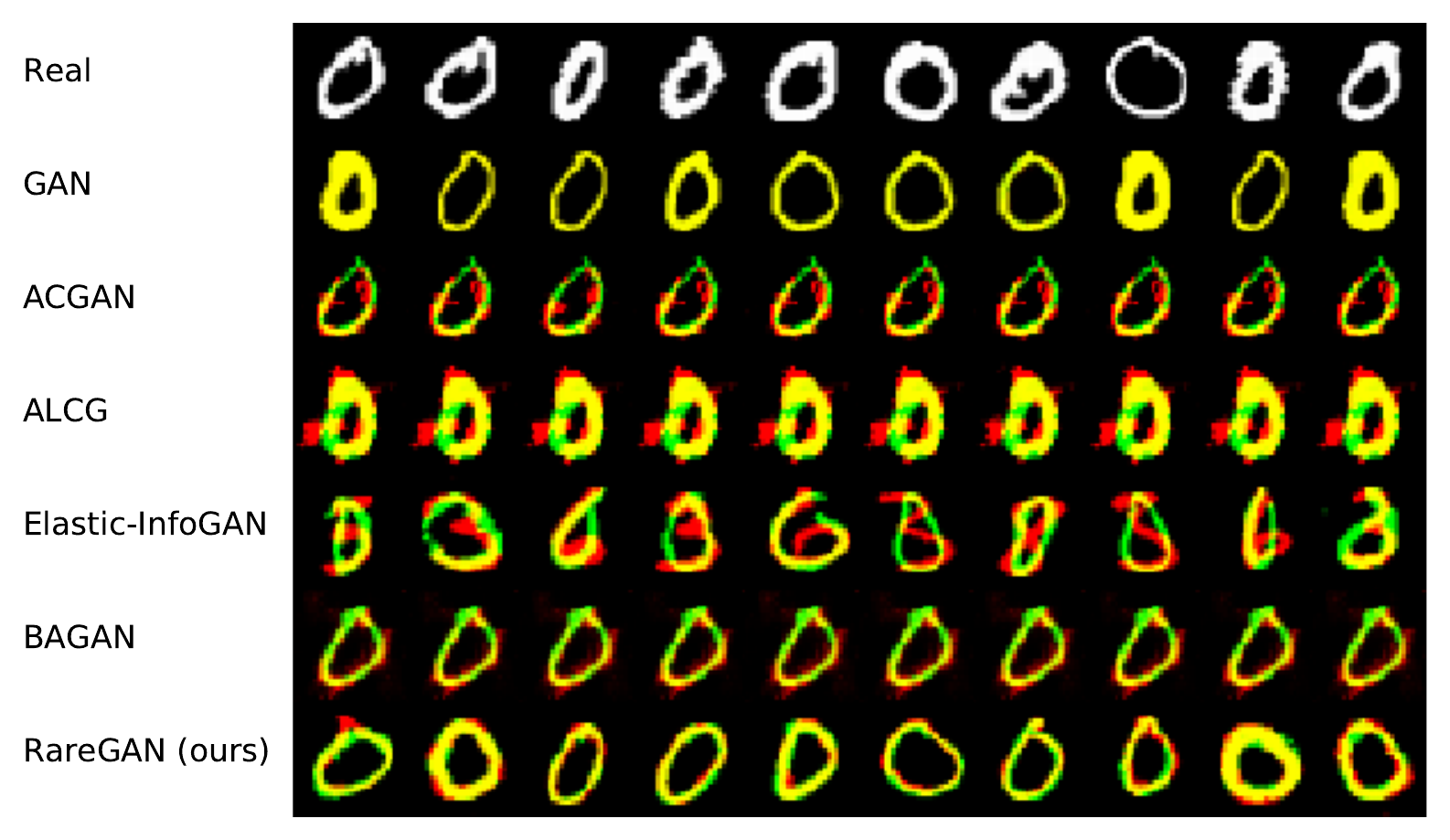}
	\vspace{-0.6cm}
	\caption{Random generated samples (no cherry-picking) on \mnist{} class '0'  with $B=\numprint{1000}$ and $\rarefrc=1\%$. 
		The red channel plots a generated image, and the green channel the nearest real image from the training set.
		Yellow pixels show where the two overlap.
		\name{} achieves the high sample quality and diversity without memorizing training data.
	}
	\label{fig:mnist_sample}
	\vspace{-0.3cm}
\end{figure}

We show that \name{} achieves a better fidelity-diversity tradeoff on the rare class than baselines
across different use cases, budgets, rare class fractions, GAN losses, and architectures.
\cref{tbl:ampmap} shows that \name{} achieves better fidelity and diversity (with a smaller labeling budget) when generating DNS amplification attack packets, compared to a state-of-the-art domain-specific technique \cite{moon2021accurately}.
Although \name{} is primarily motivated from the applications in security, networking, and systems, we also consider image generation, both as a useful tool in its own right and to visualize the improvements. %
\cref{fig:mnist_sample} shows generated samples trained on a modified \mnist{} handwritten digit dataset \cite{mnist} where we artificially forcing `0' digit as the rare class (1\% of the training data).
ACGAN \cite{odena2017conditional}, ALCG \cite{xie2019learning}, and BAGAN \cite{mariani2018bagan} produces severely mode-collapsed samples.
Elastic-InfoGAN \cite{ojha2019elastic} produces samples from the wrong class.
GAN memorizes the training dataset.
\name{} (bottom) produces high-quality, diverse samples from the correct class without memorizing the training data.

\section{Problem Formulation and  Use Cases}
\label{sec:formulation}

\myparatightest{Problem formulation}
We focus on learning a generative model for a ``rare'' (under-represented) class, subject to two constraints:
(1) We assume  a limited budget for labeling training data. 
(2) We only want to learn the rare class distribution, not the common class.  
More precisely,
we are given  i.i.d samples $\dataset=\brc{x_1,...,x_\numsample}$ from a mixture distribution $\p=\rarefrc \prare + (1-\rarefrc) \pcommon$, where $\prare$ denotes the \emph{rare} class distribution,  $\pcommon$ is  the \emph{common} class distribution, $\rarefrc\ll 1$ denotes the weight of the rare class, and we have $\support{\prare} \cap\support{\pcommon}=\emptyset$. 
Each sample $x_i$ has a label $y_i\in \{\textrm{rare, common}\}$.
The training dataset $\dataset$ does not include these labels $y_1,...,y_n$ beforehand, but we can request to label up to $\requestbgt$ samples during training. 
Given this budget $B$, we want to learn a generative model $\pgeneraterare$ that \emph{faithfully} reproduces the rare class distribution, i.e., to minimize $d\bra{\pgeneraterare, \prare}$, where $d\bra{\cdot,\cdot}$ is a %
distance metric between  distributions.

This formulation is motivated from the following use cases.

\myparatightest{Motivating scenario 1: amplification attacks (security)} 
Many widely deployed public servers and protocols like DNS, NTP, and Memcached are vulnerable to amplification attacks \cite{moon2021accurately,rossow2014amplification}, where the attacker send requests (network packets) to public servers with spoofed source IP addresses, so the response goes to the specified victims. These requests are designed to maximize response size, thus exhausting victims' bandwidth. Server operators want to know which requests trigger these \emph{amplification attacks} to e.g., drop attack requests. 
Prior solutions require detailed information about the server, such as source code \cite{rossow2014amplification}, which may be unavailable \cite{lin2019towards}. 

\emph{Our problem formulation.} As in \cite{moon2021accurately}, we treat the  \emph{rare class} as all requests with an amplification factor $\nicefrac{\bra{\textrm{size of response packet}}}{\bra{\textrm{size of request packet}}}\geq T$, a pre-defined threshold. All other requests belong to the \emph{common class}. To label a request (i.e., check its amplification factor), we send the request through the server, which can be costly. 
Hence, we want to limit the number of label queries. 
We want to learn a uniform distribution over high-amplification requests to maximize coverage of the input space.

\myparatightest{Motivating scenario 2: performance stress testing (systems \& networking)}
Many deployed systems and networks today rely on black-box components  (e.g., lacking source code, detailed specifications). 
System operators may therefore want to understand worst-case system performance (e.g., CPU/memory usage or delay in the presence of congestion) and optimize for such scenarios \cite{castan}. 
However, current tools for generating such workloads often rely on a system's source code \cite{caballero2007polyglot,petsios2017slowfuzz,castan}, which may be unavailable \cite{lin2019towards,black2021guidelines}.

\emph{Our problem formulation.}  We treat the \emph{rare class} as packets with resource usage (e.g., CPU/memory/processing delay) $\geq T$, a pre-defined threshold. The operator can use the trained $\pgeneraterare$ to synthesize such workloads. For the same reasons as the previous case, we want to limit the number of label queries and learn the rare class faithfully.

\myparatightest{Motivating scenario 3: inspecting rare class images (ML)} 
Prior GANs on unbalanced image datasets focus on generating samples from \emph{both} rare and common classes for improving downstream classification accuracy (\cref{sec:related}).  However, in some other applications, we may \emph{only} need rare class samples. For example, in federated learning, we may want to inspect the samples from specific client/class slices that have bad accuracy for debugging \cite{augenstein2019generative}.

We will see in \cref{sec:results} that \name{} outperforms baselines across all these very different use cases and data types.

%% file: tex/background.tex
\section{Background and Related Work}

\subsection{Background}
\label{sec:background}

\myparatightest{Generative Adversarial Networks (GANs)}
GANs \cite{goodfellow2014generative} are a class of deep generative models that have spurred significant interest in recent years. 
GANs involve two neural networks: a generator $G$ for mapping a random vector $z$ to a random sample $G(z)$, and a discriminator $D$ for guessing whether the input image is generated or from the real distribution $\p$. The vanilla GAN loss \cite{goodfellow2014generative}  is $ \min_G \max_D \lganjs\bra{D,\pgenerate;\p}$, where
\begin{gather}
	\scalebox{0.86}{
	$\lganjs\bra{D,\pgenerate;\p} = \Eb_{x\sim p} \brb{\log D(x)} + \Eb_{x\sim \pgenerate} \brb{\log \bra{1-D(x)}},\label{eq:gan_js_loss}$
}
\end{gather}
and $\pgenerate$ denotes the generated distribution induced by $G(z)$ where $z$ is sampled from a fixed prior distribution $p_z$ (e.g., Gaussian or uniform). It has been shown that under some assumptions, \cref{eq:gan_js_loss} is equivalent to $\min_G \djs\bra{p,\hat{p}}$, where $\djs\bra{\cdot,\cdot}$ denotes Jensen-Shannon divergence between the two distributions. Several other distance metrics have later been proposed to improve the stability of training \cite{arjovsky2017wasserstein,gulrajani2017improved,nowozin2016f,mao2017least}. Wasserstein distance $\dw\bra{\cdot,\cdot}$ is one of the most widely used metrics \cite{arjovsky2017wasserstein,gulrajani2017improved}. The loss of Wasserstein GAN is $\min_G \dw\bra{\p,\pgenerate}= \min_G \max_{\brl{D}\leq 1} \lganw\bra{D,\pgenerate;\p}$, where
\begin{align}
	 \lganw\bra{D,\pgenerate;\p} = \Eb_{x\sim p} \brb{ D(x)} - \Eb_{x\sim \pgenerate} \brb{ D(x)}
	\label{eq:gan_w_loss}
\end{align}
and $\brl{D}$ denotes the Lipschitz constant of $D$. \name{} works well with both of these losses.

\myparatightest{Auxiliary Classifier GANs (ACGAN)}
\label{sec:acgan}
Conditional GANs (CGANs) \cite{mirza2014conditional,odena2017conditional}  are a variant of GANs that support conditional sampling. Besides $z$, the generators in CGANs have an additional input $c$ which controls the properties (e.g., category) of the generated sample. For example, in a face image dataset (with male/female labeled), instead of only sampling from the entire face distribution, generators in CGANs could allow us to control whether to generate a male or female by specifying $c$. Several different techniques have been proposed to train such a conditional generator \cite{mirza2014conditional,odena2017conditional,salimans2016improved,mariani2018bagan}.  ACGAN \cite{odena2017conditional} is one such widely-used variant \cite{kong2019active,xie2019learning,choi2018stargan,liang2020controlled}. ACGAN adds a classifier $C$ which discriminates the labels for both generated data and real data. The ACGAN loss function is:
\begin{gather}
	\scalebox{0.85}{
	$\min_C \min_G \max_D \lgan\bra{D,\pgenerate;\p} + \lclassification\bra{C,\pgeneratesamplelabel;\psamplelabel},
	\label{eq:acgan_loss}$
}
\end{gather}
where  $\lgan(D,\pgenerate;\p)$ is the regular GAN loss (e.g.,  \cref{eq:gan_js_loss} \cite{kong2019active,xie2019learning,choi2018stargan} or \cref{eq:gan_w_loss} \cite{acgangithub}), except that 
	$\pgenerate$  is induced by $G(z,c)$, where $z\sim p_z$ and $c\sim \plabel$,
where  $\plabel$ is the ground truth label distribution. $\lclassification$ is defined by
\begin{align}
	&\scalebox{0.9}{
	$\lclassification\bra{C,\pgeneratesamplelabel;\psamplelabel} =$}\nonumber\\
	&\scalebox{0.9}{
	$-\Eb_{\bra{x,c}\sim \psamplelabel} \brb{\log C(x,c)} - \Eb_{(x,c)\sim \pgeneratesamplelabel} \brb{\log C(x,c)} ,
	\label{eq:classification_loss}$
}
\end{align}
where $C(x,c)$ denotes classifier $C$'s probability prediction for class $c$ on input sample $x$, $\psamplelabel$ denotes the  real joint distribution of samples and labels, and $\pgeneratesamplelabel$ denotes the joint distribution over labels and generated samples in $\pgenerate$. 
In practice, $D$ and $C$ usually share some layers.  Note that in \cref{eq:classification_loss} the classifier is trained to match not only the real data, but also the generated data, as in \cite{kong2019active,acgangithub,odena2017conditional}. However, in some other implementations, the second part of loss only applies on $G$, so that the classifier will not be misled by errors in the generator \cite{xie2019learning,choi2018stargan}.

\subsection{Related work}
\label{sec:related}
Depending on the availability of the labels, prior related works can be classified into fully-labeled, unsupervised, semi-supervised, and  self-supervised GANs.

\myparatightest{Fully supervised GANs}
Prior work has studied how to use GANs (particularly ACGAN) to augment imbalanced,  \emph{labeled} datasets, e.g.,  for  downstream classification tasks~\cite{mullick2019generative,douzas2018effective,ren2019ewgan,ali2019mfc,mariani2018bagan,rangwani2021class,asokan2020teaching,yang2021ida}. 
For example, EWGAN \cite{ren2019ewgan}, MFC-GAN \cite{ali2019mfc}, \citet{douzas2018effective},  \citet{wei2019generative}, and BAGAN \cite{mariani2018bagan} all augment the original dataset by generating samples from the minority class with a conditional GAN.
\citet{wei2019generative} utilizes known \emph{mappings} between images in different classes (e.g., mapping an image of normal colon tissue to precancerous colon tissue).
BAGAN \cite{mariani2018bagan} instead trains an autoencoder on the entire dataset, learns a Gaussian latent distribution for each class, and uses that as the input noise for each class to the GAN generator. 
We cannot utilize these approaches because we lack labels.

\myparatightest{Unsupervised GANs}  Unsupervised GANs \cite{chen2016infogan,ojha2019elastic,lin2020infogan} do not control which factors to learn. Hence, there is no guarantee that they will learn to separate samples along the desired factor and threshold (e.g., classification time of the generated packets). 

\myparatightest{Semi-supervised GANs} Our proposed approach in \cref{sec:approach_unlabeled}  is one instance of semi-supervised GANs \cite{odena2016semi,salimans2016improved,dai2017good,kumar2017semi,haque2020ec,zhou2018gan}. 
Other semi-supervised GANs could also be used, like the seminal one \cite{salimans2016improved}, which uses a single modified discriminator both to separate fake from real samples (as in classical GANs) and to classify the labels of real data. We choose to use separated classifier (as in ACGAN) as the classifier is not influenced by real/fake objective and  therefore provides cleaner signal for our active learning technique in \cref{sec:approach_active}. 
The closest prior work, \textbf{ALCG \cite{xie2019learning}}, is also an instance of semi-supervised GANs.
Like us, they are training conditional GANs in an active learning setting. 
However, their goal is to synthesize high-quality samples from \emph{all} classes, whereas we want to faithfully reproduce only the rare class. 
We show how this distinction requires different %
 algorithmic designs
 (\cref{sec:approach}), and leads to poor performance by ALCG on our problems (\cref{sec:exp}).

\myparatightest{Self-supervised GANs} Self-supervision has been used in both unsupervised and semi-supervised GANs \cite{sun2020matchgan,chen2019self,ojha2019elastic}.
It is also unclear how to apply self-supervised GANs in our problem, as they are most useful when we have some prior understanding of the physical or semantic of a system, but in our problem we are given an arbitrary system whose internal structure is unknown. For example, for disentangling digit types (e.g., 0 v.s. 1), Elastic-InfoGAN \cite{ojha2019elastic} applies operations like rotation on images to construct positive pairs for self-supervised loss, as we know these operations do not change digit types. However, it is unclear what corresponding operations should be in our problem due to the `black box' nature of the systems. For example, in motivating scenario 2 (\cref{sec:formulation}), it is unclear what operations on packets would keep the CPU/memory usage of the system.

Our work is also related to prior work on weighted loss and data augmentation for GANs, which we discuss below.
\myparatightest{Weighted loss} Our proposed approach involves a weighted loss for GANs (\cref{sec:approach_weight}). Although weighted loss for GANs and classification has been proposed in prior work \cite{zadorozhnyy2021adaptive, cui2019class}, the weighting schemes and the goals are completely different to ours. For example, awGAN \cite{zadorozhnyy2021adaptive} adjusts the weights on the real/fake losses, with the goal of balancing the gradient directions of these two losses and making training more stable. Instead, as we see in \cref{sec:approach_weight}, the proposed \name{} uses different weights for samples in the rare and the common classes (regardless of whether they are fake or real), with the goal of balancing the learning of the rare and the common classes.

\myparatightest{Data augmentation} Data augmentation techniques have been proposed for improving GANs' performance in limited data regimes \cite{karras2020training, zhao2020differentiable}. However, these techniques usually focus on image datasets and the augmentation operations (e.g., random cropping) do not extend to other domains (e.g., our networking datasets).

%% file: tex/approach.tex
\section{Approach}
\label{sec:approach}
As mentioned in \cref{sec:formulation} and  \cref{sec:related}, our problem is distinguished by two factors:
(1) We want to learn \emph{only} the rare class distribution; (2) The rare/common labels are \emph{not} available in advance, and we have a fixed labeling budget (can be used online during training).

\label{sec:approach_naive}
The most obvious straw-man solution to our problem is to randomly and uniformly draw $\requestbgt$ samples from the dataset and request their labels. Then, we train a \emph{vanilla GAN} on the packets with label ``rare''. However, since the rare class could have a very low fraction, the number of training samples will be small and the GAN is likely to overfit to the training dataset and generalize poorly.

As in prior work \cite{xie2019learning,ren2019ewgan,ali2019mfc,mariani2018bagan}, we can use %
conditional GANs like ACGAN (\cref{sec:background}) to incorporate common class samples into training, because they could actually be useful for learning the rare class. For example, in face image datasets, the rare class (e.g., men with long hair) and common classes share same characteristics (i.e., faces). 
However, due to the small number of rare samples, ACGAN still has bad fidelity and diversity (\cref{sec:results}). 

In the following, we progressively discuss the design choices we make in \name{} to address the challenges,
and highlight the differences to ALCG (\cref{sec:related}), the most closely-related work.

\subsection{Better distribution learning with unlabeled samples}
\label{sec:approach_unlabeled}
In the above process, the majority of samples from $\dataset$ are unlabeled because of the labeling budget. Those samples are not used in training (as in ALCG). However, they  contain information about the mixture distribution of rare and common classes, and could therefore help  learn  the rare class. 

Our proposed approach relies on carefully altering the ACGAN loss.
Recall that the loss has two parts: a classification loss separating rare and common samples, and a GAN loss evaluating their mixture distribution (\cref{sec:acgan}). Note that the GAN loss does \emph{not} require labels. We  propose a modified ACGAN training that uses labeled samples for the classification loss, and \emph{all} samples for the GAN loss.
However, when training the GAN loss, we need to know the \emph{fraction} of rare/common classes in order to feed the condition input to the generator.
This can be estimated from the labeled samples.
The maximum likelihood estimate $\hat{\alpha}=\frac{x}{n}$ has variance  $\frac{\rarefrc(1-\rarefrc)}{n}$, which is small for  reasonable $n$.

\subsection{Improving classifier performance with active learning}
\label{sec:approach_active}
Because the rare and common classes are highly imbalanced, the classifier in ACGAN could have a bad accuracy. In classification literature, confidence-based active learning has been widely used for solving this challenge \cite{li2006confidence,joshi2009multi,sivaraman2010general}, which expends the labeling budget on samples about which the classifier is least confident.

Inspired by these works, our approach is to divide the training into $\numstage$ stages. At the beginning of each stage, we pass all unlabeled samples through the classifier, and request the labels for the  $\nicefrac{\requestbgt}{\numstage}$ samples that have the lowest $\max\brc{ C(x,\textrm{rare}), C(x,\textrm{common})}$, where $C(x,c)$ denotes the classifier's (normalized) output.\footnote{In the first stage, the samples for labeling are randomly chosen from the dataset.} 
This sample selection criterion is called ``least confidence sampling'' in prior literature \cite{lewis1994heterogeneous}. There are other criteria like margin of confidence sampling and entropy-based sampling \cite{vlachos2008stopping,kong2019active,xie2019learning}. Since we have only two classes, they are in fact equivalent.

While ALCG also uses confidence-based active learning, %
it does so in a diametrically opposite way:
they request the labels for the \emph{most certain} samples. These two completely different designs result from having different goals: ALCG aims to generate high-quality images, whereas we aim to faithfully reproduce the rare class distribution. The most certain samples usually have better image quality, and therefore ALCG wants to include them in the training. The least certain samples are more informative for learning distribution boundaries, and therefore we label them.

\myparatightest{Relation to \cref{sec:approach_unlabeled}}
Naively using active learning is actually counterproductive;
if we only use labeled samples in the training (as is done in ALCG), the learned rare distribution will be biased, which partially explains ALCG's poor performance in our setting (\cref{sec:exp}). 
As discussed in \cref{sec:approach_unlabeled}, we instead use all unlabeled samples for training the GAN loss.
The following proposition explains how this circumvents the problem (proof in \cref{app:proof_unlabeled}). %
\begin{proposition}
\label{thm:unlabeled}
The optimization
    $$
    \p^* \in \arg\min_{\pgenerate} \min_{C\in \mathcal C} d\bra{\pgenerate,\p} + \lclassification\bra{C,\pgeneratesamplelabel;\psamplelabel'}
    $$ 
    satisfies $d\bra{\p_r^*,\p_r}=0$, where: (a) $\p_r^*$ is $\p^*$ under condition ``rare'', (b) $\psamplelabel'$ is \textbf{any} joint (sample, label) distribution where the support of $\psamplelabel'$ covers the entire sample space, (c) $\pgeneratesamplelabel$ is the generated joint distribution of  samples and labels, and (d) $\mathcal C$ is the set of measurable functions.
\end{proposition}

The above optimization is a generalization of the ACGAN loss function, where $d(\cdot, \cdot)$ denotes an appropriately-chosen distance function for the GAN loss (e.g., Wasserstein-1). Note that the classification loss is computed over the (possibly biased) distribution induced by active learning, $p'$, whereas the GAN loss is computed with respect to the true distribution $p$, which uses all samples, labeled or not. The proposition is saying that even though we use the biased $p'$ for the classification loss, we can still learn $p_r$. On the other hand, if we had used $p'$ for the GAN loss (as is done in ALCG), we would recover a biased version of $p_r$.

\subsection{Better rare class learning with weighted loss}
\label{sec:approach_weight}
Because the rare class has low mass, errors in the rare distribution have only bounded effect on the GAN loss in \cref{eq:acgan_loss}. 
Next, we propose a re-weighting technique for reducing this effect at the expense of learning the common class.
Let $\pgenerate$ be the learned sample mixture distribution (without labels). 
Let $\rarefrclearned=\int_{\support{\prare}}\pgenerate(x)dx$ be the fraction of rare samples under $\pgenerate$, 
and let $\pgenerateraresupport$ be $\pgenerate$ restricted to and normalized over $\support{\prare}$. (Recall that $\pgeneraterare$ is the generated distribution under condition $y=$``rare''; it need not be the case that $\support{\prare}=\support{\pgeneraterare}$.)
Similarly, let $\pgeneratecommonsupport$ be $\pgenerate$ over $\mathcal X-\support{\prare}$ where $\mathcal X$ is the entire sample space.
Recall that the original GAN loss in \cref{eq:acgan_loss} tries to minimize 
$$
d\bra{\p,\pgenerate}=d\bra{\rarefrc \cdot \prare + (1-\rarefrc) \cdot \pcommon,~ \rarefrclearned \cdot \pgenerateraresupport + (1-\rarefrclearned) \cdot \pgeneratecommonsupport}
$$ 
where $d$ is $\djs$ or $\dw$.  We propose to modify this objective function to  instead minimize
\begin{gather}
	\scalebox{0.88}{
	$d\bra{\weight\rarefrc \cdot \prare + \cdot(1-\weight\rarefrc) \cdot \pcommon,~ 
	\frac{\weight\rarefrclearned}{s} \cdot  \pgenerateraresupport +  \frac{\bra{1-\weight \rarefrc}\cdot \bra{1-\rarefrclearned}}{s\cdot\bra{1-\rarefrc}}\cdot \pgeneratecommonsupport}, \label{eq:weighted_distance}$}
\end{gather}
where $\weight\in\bra{1,1/\alpha}$ is the additional multiplicative  weight to put on the rare class; and $s=\weight\rarefrclearned + \frac{1-\weight \rarefrc}{1-\rarefrc} \bra{1-\rarefrclearned}$ is the normalization constant, which is $1$ when $\rarefrc=\rarefrclearned$. It is straightforward to see that \cref{eq:weighted_distance}$ =0 \Leftrightarrow d\bra{\p,\pgenerate}=0$. However, these two objective functions have different effects in training: this modified loss will more heavily penalize errors in the rare distribution. Consider two extremes: (1) When $w=1$, \cref{eq:weighted_distance} is reduced to the original $d\bra{\p,\pgenerate}$, placing no additional emphasis on the rare class; (2) When $w=\nicefrac{1}{\rarefrc}$, \cref{eq:weighted_distance} is reduced to $d\bra{\prare,\pgenerateraresupport}$,  focusing only on the rare class and placing no constraint on the common class. This  completely loses the benefit of learning the classes jointly (\cref{sec:approach_naive}). For a $w\in\bra{1,1/\alpha}$, we can achieve a better trade-off between the information from the common class and the penalty on the error of rare class.

To implement the above idea, we propose to add a multiplicative weight to the loss of both real and generated samples according to their label, i.e. changing \cref{eq:gan_js_loss,eq:gan_w_loss} to 
\begin{align}
	&\scalebox{0.86}{
	$\lganjsweighted\bra{D,\pgenerate;\p} =$}\nonumber\\
	&\scalebox{0.86}{
	$\Eb_{x\sim \p} \brb{W(x)\cdot \log D(x)} + \frac{1}{s} \cdot\Eb_{x\sim \pgenerate} \brb{W(x)\cdot\log \bra{1-D(x)}}
	\label{eq:gan_js_loss_weighted}$}
\end{align}
for Jensen-Shannon divergence and
\begin{align}
	 &\scalebox{0.86}{$\lganwweighted\bra{D,\pgenerate;\p} =$}\nonumber\\
	 &\scalebox{0.86}{$ \Eb_{x\sim \p} \brb{ W(x)\cdot D(x)} -  \frac{1}{s}\cdot \Eb_{x\sim \pgenerate} \brb{ W(x) \cdot D(x)}\label{eq:gan_w_loss_weighted}$}
\end{align}
for Wasserstein distance, where
\begin{align}
	W(x)=\left\{\begin{matrix}
		\weight&(x\in \support{p_r})\\
		\frac{1-\weight\rarefrc}{1-\rarefrc}&(x \not\in \support{p_r})
	\end{matrix}\right.\;.
\label{eq:weight}
\end{align}
Using \cref{eq:gan_js_loss_weighted} and \cref{eq:gan_w_loss_weighted} is equivalent to minimizing  \cref{eq:weighted_distance} for $\djs$ and $\dw$ respectively (proof in \cref{app:proof_weight}): 
\begin{proposition}\label{thm:weight}
	For any $D$, $\p$, and $\pgenerate$, we have 
	\begin{align}
		\lganjsweighted\bra{D,\pgenerate;\p} = \lganjs\bra{D,\qgenerate;\q}\\
		\lganwweighted\bra{D,\pgenerate;\p} = \lganw\bra{D,\qgenerate;\q}
	\end{align}
where $\qgenerate=\frac{\weight\rarefrclearned}{s} \cdot  \pgenerateraresupport +  \frac{\bra{1-\weight \rarefrc}\cdot \bra{1-\rarefrclearned}}{s\cdot\bra{1-\rarefrc}}\cdot \pgeneratecommonsupport$, and $\q=\weight\rarefrc \cdot \prare + \cdot(1-\weight\rarefrc) \cdot \pcommon$.
\end{proposition}

Implementing this weighting is nontrivial, however: (1) The above implementation requires the ground truth labels of all real and generated samples for evaluating \cref{eq:weight}, and we do not want to waste labeling budget on weight estimation. For this, we use the ACGAN rare/common classifier as a surrogate labeler. Although this classifier is imperfect, weighting real and generated samples according to the same labeler is sufficient to ensure that the optimum is still $d\bra{\p,\pgenerate}=0$. (2) Evaluating the normalization constant $s$ in \cref{eq:gan_w_loss_weighted,eq:gan_js_loss_weighted} requires estimating $\rarefrclearned$, which  is inefficient as $\rarefrclearned$  changes during training. Empirically, we found that setting $s=1$ gave good and stable results.

%% file: tex/experiments.tex
\section{Experiments}
\label{sec:exp}
We conduct experiments on all three applications in \cref{sec:intro}. The code can be found at \url{https://github.com/fjxmlzn/RareGAN}.

\myparatightest{Use case 1: DNS amplification attacks} 
DNS is one of the most widely-used protocols in amplification attacks \cite{rossow2014amplification}. DNS requests that trigger high amplification have been extensively analyzed in the security community \cite{kambourakis2007fair,anagnostopoulos2013dns,rossow2014amplification}, though most of the those analyses are manual or use tools  specifically designed for (DNS) amplification attacks.  We show that \name{}, though designed for a more general set of problems, can also be effectively used for finding amplification attack requests.
In this setting, we define the rare class as DNS requests that have $\frac{\text{size~of~response}}{\text{size~of~request}}\geq T$, where $T$ is a threshold specified by users. For the request space, we follow  the configuration of \cite{lin2019towards,moon2021accurately}: we let GANs generate 17 fields in the DNS request;  for 5 fields among them (\textit{qr}, \textit{opcode}, \textit{rdatatype}, \textit{rdataclass}, and \textit{url}), we provide candidate values; 
for all other 12 fields, we let GANs explore all possible bits. %
The entire search space is $3.6\times 10^{17}$. 
Unlike image datasets where samples from the mixture distribution $\p$ are  given, here we need to define $\p$. Since our goal is to find all DNS requests with amplification $\geq T$, we define $\p$ as a uniform distribution over the search space.  More details are in \cref{app:experiment_details}. 

\emph{Note on ethics:} For this experiment, we needed to make many DNS queries. To avoid harming the public DNS resolvers, we set up our own DNS resolvers on Cloudlab \cite{duplyakin2019design} for the experiments. \footnote{These experiments did not involve collecting any sensitive data.  Such ``penetration testing'' of services is common practice in the security literature and we followed best practices~\cite{ethicssecurity}. Two leading guidelines are \emph{responsible disclosure} and \emph{avoid unintentional harm}. We avoided harming the public Internet by running our experiments in sandboxed environments. Since we only reproduced synthetic (known) attack modes~\cite{moon2021accurately}, we did not need to disclose new vulnerabilities.}

\myparatightest{Use case 2: packet classification}  Network packet classification is a fundamental building block of modern networks. Switches or routers classify incoming packets to determine what action to take (e.g., forward, drop) \cite{liang2019neural}. 
An active research area in networking is to propose classifiers with low inference latency \cite{chiu2018design,liang2019neural,soylu2020bit,rashelbach2020computational}.
We take a recently-proposed packet classifier \cite{liang2019neural} for example, which was designed to optimize classification time and memory footprint. 
We define the rare class as network packets that have $\text{classification time}\geq T$, a threshold specified by users. GANs generate the bits of 5 fields: \textit{source/destination IP}, \textit{source/destination port}, and \textit{protocol}. The search space is $1.0\times 10^{31}$. As before, $\p$ is a uniform distribution over the entire search space.
To avoid harming network users, we ran all measurements on our own infrastructure rather than active switches. 
More details are in \cref{app:experiment_details}. 

\myparatightest{Use case 3: inspecting rare images}%
Although \name{} is primarily designed for the above use cases, we also use images for visualizing the improvements.
Following the settings of related work \cite{mariani2018bagan}, we simulate the imbalanced dataset with widely-used datasets: \mnist{} \cite{mnist} and \cifar{} \cite{cifar}. For \mnist{}, we treat digit 0 as the rare class, and all other digits as the common class. For \cifar{}, we treat airplane as the rare class, and all other images as the common class. In both cases, the default class fraction is 10\%. To simulate a smaller rare class, we randomly drop images from the rare class.

\subsection{Evaluation Setup}
\label{sec:setup}

\myparatightest{Baselines}
To demonstrate the effect of each design choice, we compare all intermediate versions of \name{}: vanilla GAN (\cref{sec:approach_naive}), ACGAN (\cref{sec:approach_naive}), ACGAN trained with all unlabled samples (\cref{sec:approach_unlabeled}), plus active learning (\cref{sec:approach_active}), plus weighted loss (\cref{sec:approach_weight}). In all figures and tables, they are called: ``GAN'', ``ACGAN'', ``\name{} (no AL)'' , ``\name{}'' annotated with 1.0, and ``\name{}'' annotated with weight ($>1.0$), respectively. 
All the above baselines and \name{} use the same network architectures.
For the first two applications, the generators and discriminators are MLPs.
The GAN loss is Wasserstein distance (\cref{eq:gan_w_loss}), as it is known to be more stable than Jensen-Shannon divergence on categorical variables \cite{arjovsky2017wasserstein,gulrajani2017improved}. For the  image datasets, we follow the popular public ACGAN implementation \cite{acgangithub}, where the generator and discriminator are CNNs, and the GAN loss is Jensen-Shannon divergence (\cref{eq:gan_js_loss}). %

We also evaluate representative prior work on three directions: (1) \emph{GANs with active learning}: ALCG (using only labeled samples in training and using the most certain samples for labeling) \cite{xie2019learning}; (2) \emph{GANs for imbalanced datasets}: BAGAN \cite{mariani2018bagan};  (3) \emph{Unsupervised/Self-supervised GANs}: Elastic-InfoGAN \cite{ojha2019elastic}. We only evaluate the last two on MNIST, as they only released codes for that. As discussed in \cref{sec:related}, we cannot directly apply the last two in our problem; we make minimal modifications to make them suitable (see \cref{app:experiment_details}).

\myparatightest{Metrics}
We aim to minimize the distance between real and generated rare class $d\bra{\pgeneraterare, \prare}$. 
In practice, generative models are often evaluated along two axes: \emph{fidelity} and \emph{diversity} \cite{naeem2020reliable}.
Because the data types differ across applications, we  use different ways to quantify them:%

\myparatightest{\emph{Network packets (use cases 1, 2)}} \emph{(1) Fidelity.} Network packets are a high-dimensional list of categorical variables. We lack the ground truth $\prare$ to estimate fidelity. 
Instead, we  estimate the true distribution of ``scores" in $\prare$
(i.e., $\frac{\text{size~of~response}}{\text{size~of~request}}$ in DNS amplification attacks, and classification time in packet classifiers). 
This surrogate distribution is operationally meaningful, e.g., for quantifying mean or maximum security risk.  We define $\hrare$ as the ground truth distribution of this number over the rare class (estimated by drawing random samples from the entire search space, computing their scores, and then filtering out the scores that belong to the rare class), and $\hgeneraterare$ as its corresponding generated distribution. We use $\dwone\bra{\hrare,\hgeneraterare}$ as the fidelity metric, where $\dwone\bra{\cdot,\cdot}$ denotes Wasserstein-1 distance, as it has a simple, interpretable geometric meaning (integrated absolute error between the 2 CDFs). \emph{(2) Diversity.} When GANs overfit, many generated packets are duplicates. Therefore, we count the fraction of unique rare packets (i.e., those with a threshold score $\geq T$) in a set of 500,000 generated samples as the diversity metric.

\myparatightest{\emph{Images (use case 3)}} \emph{(1) Fidelity.} We use widely-used Fr\'{e}chet Inception Distance (FID) \cite{heusel2017gans} between generated data and real rare data to measure fidelity.  \emph{(2) Diversity.} The previous diversity metric is not applicable here, as duplicate images are very rare. Instead, we take a widely used heuristic \cite{wang2016ensembles} to check if GAN overfits to the training data: for each generated image, we find its nearest neighbor (in L2 pixel distance) in the training dataset. 
We then compute the average of nearest distances among a set of generated samples. Note that these two metrics are not completely decoupled: when GAN overfits severely, FID also detects that. 
Nonetheless, these metrics are widely used in the literature \cite{wang2016ensembles,heusel2017gans,arora2017gans,shmelkov2018good}.

\subsection{Results}
\label{sec:results}

\begin{figure}
	\centering
	\includegraphics[width=0.9\linewidth]{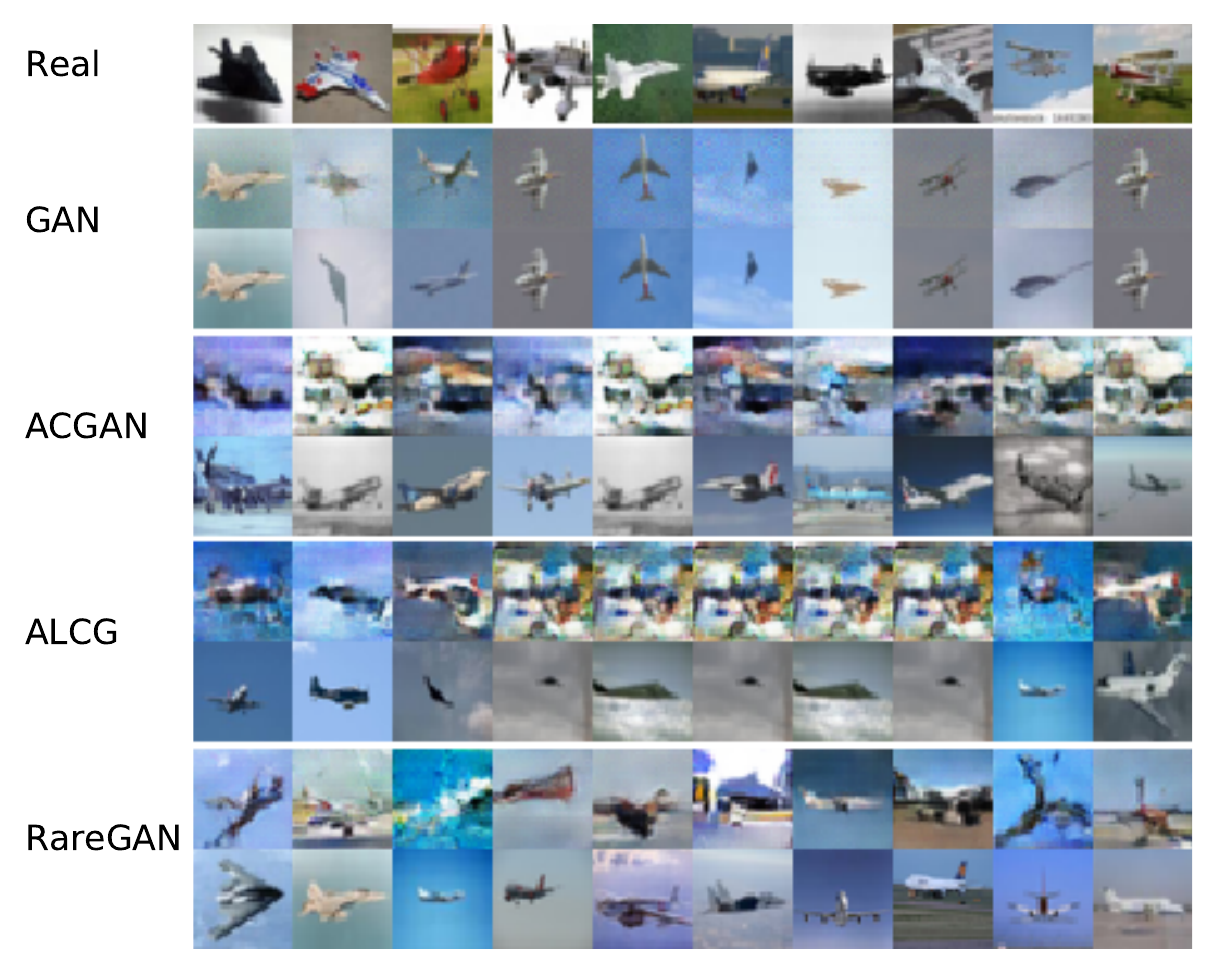}
	\vspace{-0.3cm}
	\caption{Generated samples (no cherry-picking) on \cifar{} `airplanes' with $B=\numprint{10000}$ and $\rarefrc=10\%$. Each baseline's upper row is generated samples; lower row is the closest real sample.}
	\label{fig:cifar_sample}
	\vspace{-0.3cm}
\end{figure}

Unless otherwise specified, the default configurations are: the number of stages $\numstage=2$ (for \name{} and ALCG), weight $\weight=3$ (for \name{}); in DNS, labeling budget $B=\numprint{200000}$, rare class fraction $\rarefrc=0.776\%$ (corresponding to $T=10$); in packet classification, $B=\numprint{200000}$, $\rarefrc=1.150\%$ (corresponding to $T=0.055$); in \mnist{}, $B=5,000$, $\rarefrc=1\%$; in \cifar{}, $B=\numprint{10000}$, $\rarefrc=10\%$. Note that the choice of these default configurations do not influence the ranking of different algorithms too much, as we will show in the studies later. All experiments are run over 5 random seeds. More details are in \cref{app:experiment_details}.

\begin{figure*}[!tb]
	\centering
	\begin{minipage}{0.66\textwidth}
		\begin{minipage}{0.48\linewidth}
			\centering
			\includegraphics[width=\linewidth]{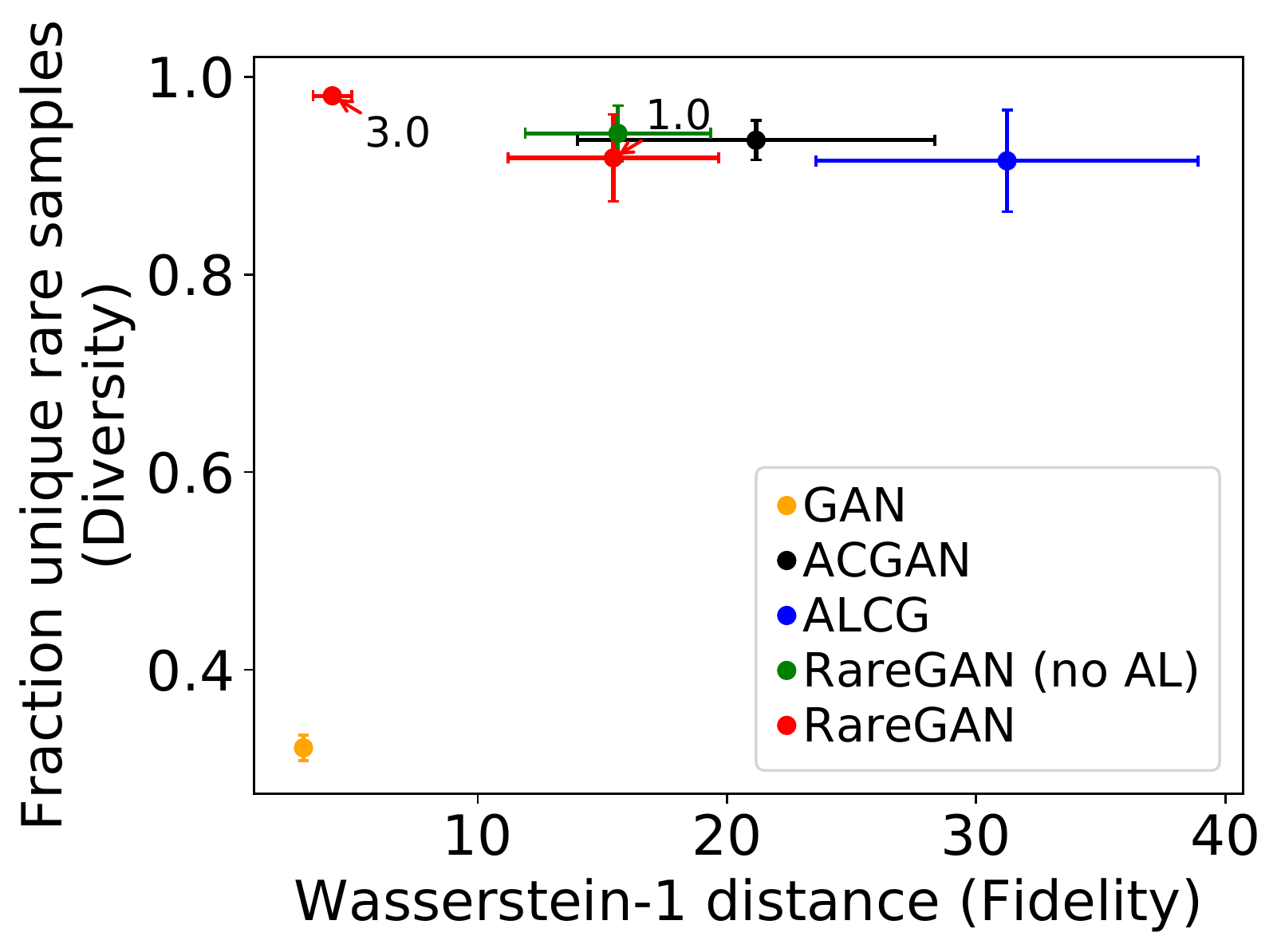}
			\vspace{-0.5cm}
			\subcaption{DNS amplification attacks.}
			\label{fig:default_dns}
		\end{minipage}~%
		\begin{minipage}{0.48\linewidth}
			\centering
			\includegraphics[width=\linewidth]{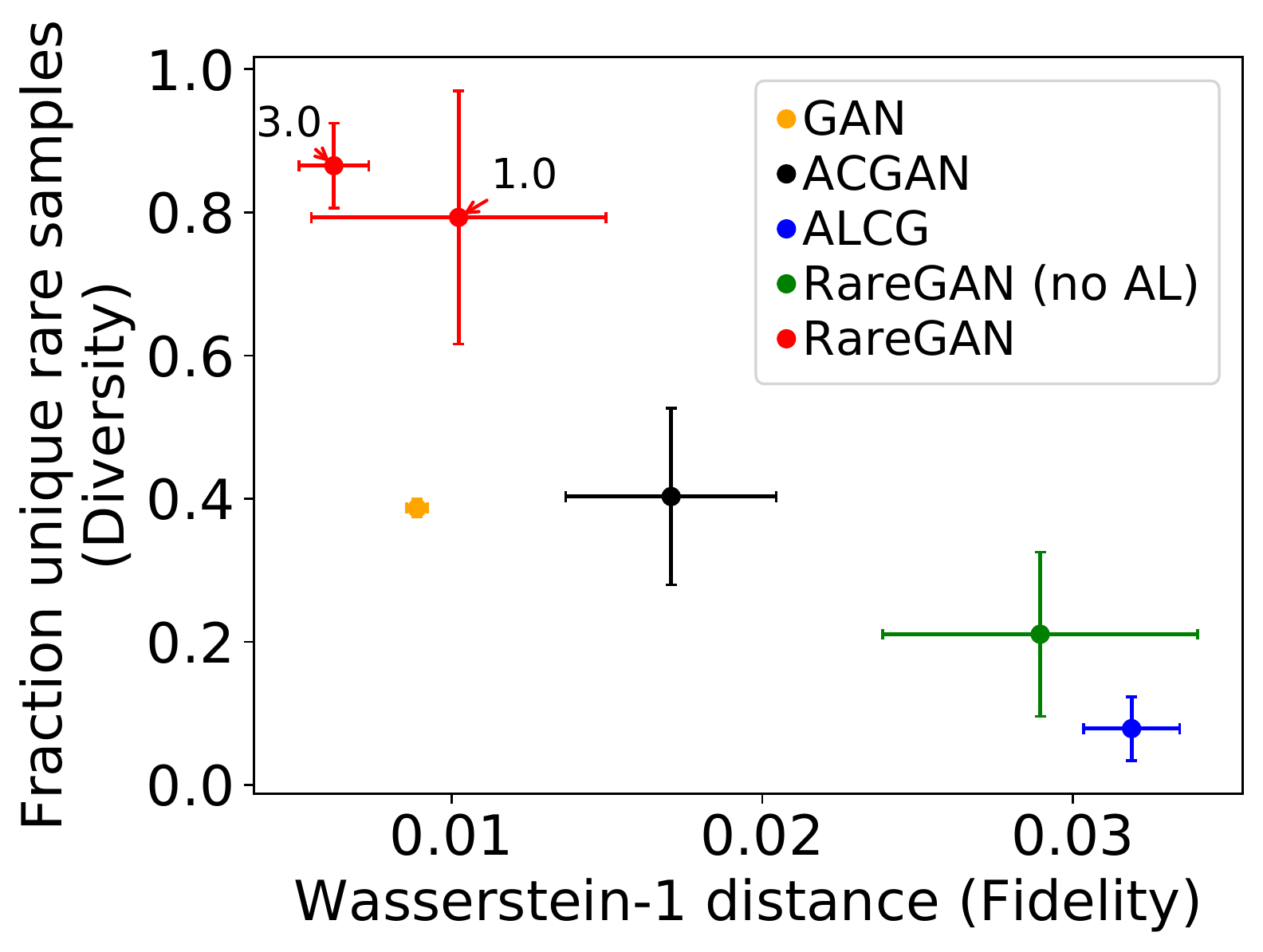}
			\vspace{-0.5cm}
			\subcaption{Packet classifiers.}
			\label{fig:default_pc}
		\end{minipage}
		\begin{minipage}{0.48\linewidth}
			\centering
			\includegraphics[width=\linewidth]{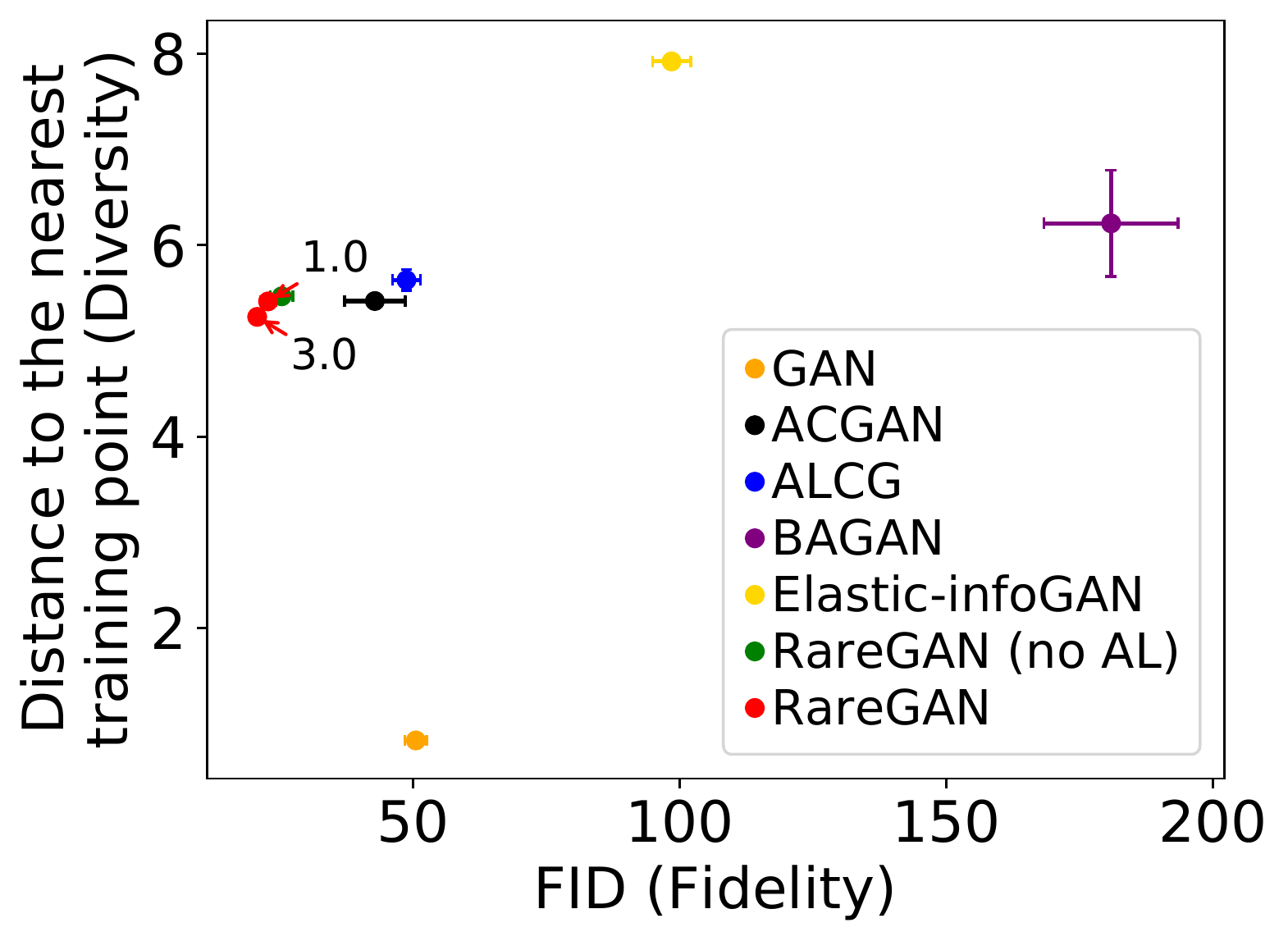}
			\vspace{-0.5cm}
			\subcaption{\mnist{}.}
			\label{fig:default_mnist}
		\end{minipage}~
		\begin{minipage}{0.48\linewidth}
			\centering
			\includegraphics[width=\linewidth]{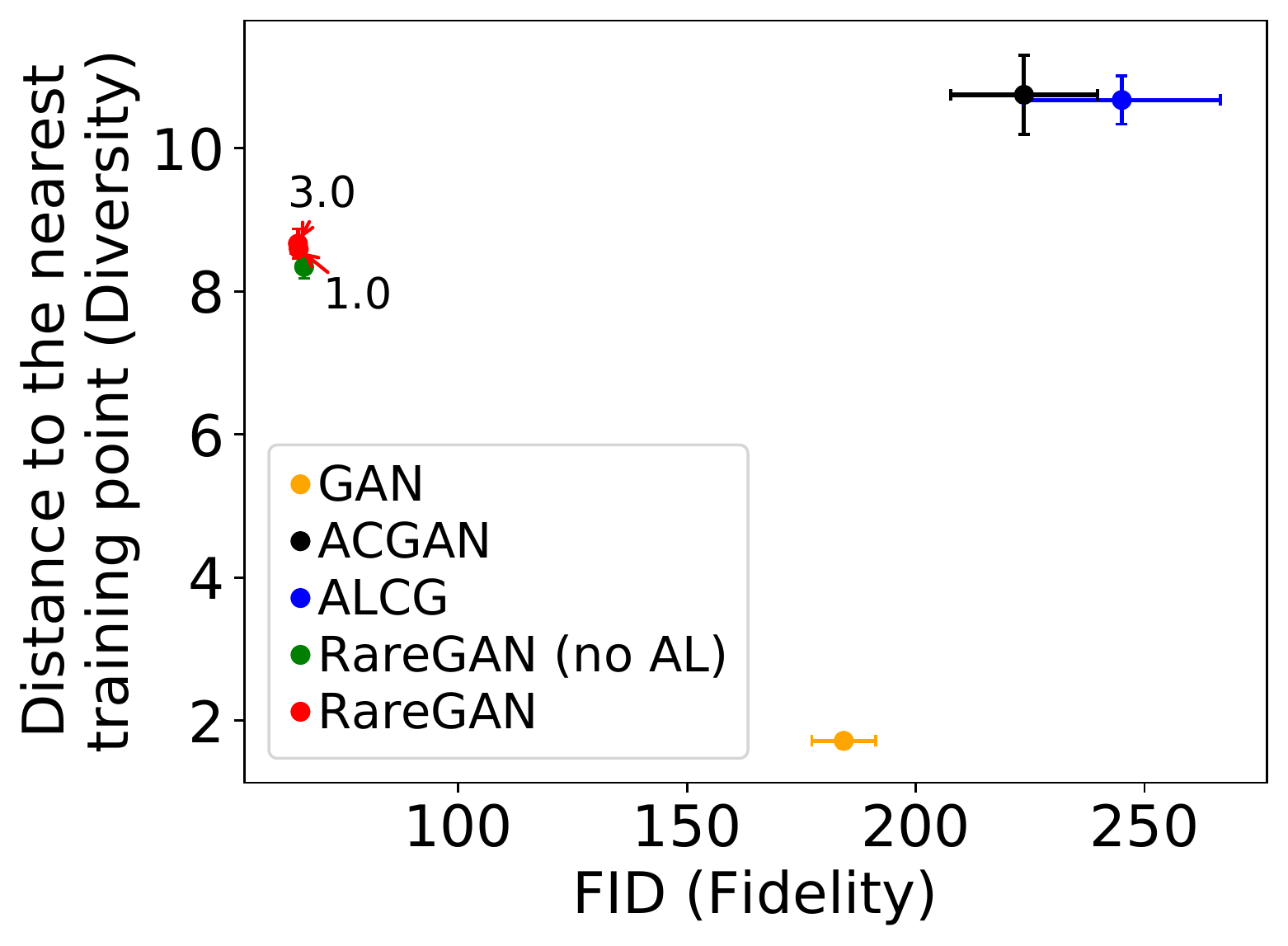}
			\vspace{-0.5cm}
			\subcaption{\cifar{}.}
			\label{fig:default_cifar}
		\end{minipage}
		\vspace{-0.2cm}
		\caption{\name{} achieves the best fidelity-diversity tradeoff on all applications. Horizontal axis is fidelity (lower is better). 
		Vertical axis is diversity (higher is better). Fidelity/diversity metrics are explained in \cref{sec:setup}. Bars show standard error over 5 runs. }%
		\label{fig:default}
	\end{minipage}%
	~~~
	\begin{minipage}{0.34\textwidth}
		\begin{minipage}{0.92\linewidth}
			\centering
			\includegraphics[width=\linewidth]{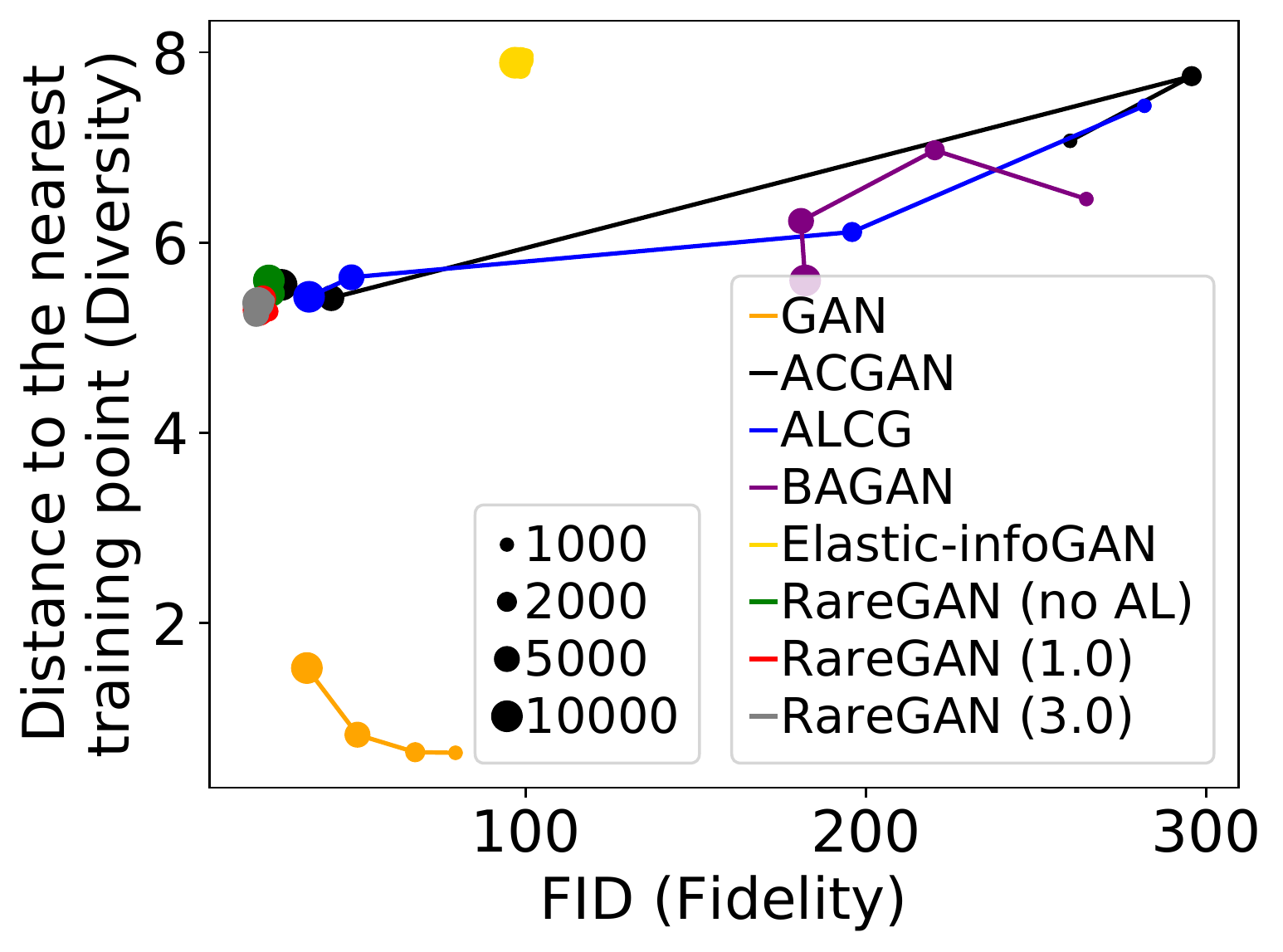}
			\vspace{-0.5cm}
			\caption{\mnist{} with different labeling budget $B$. \name{} is insensitive to $B$.}
			\label{fig:low_bgt_mnist}
		\end{minipage}%
		~~
		
		\begin{minipage}{0.92\linewidth}
			\centering
			\includegraphics[width=\linewidth]{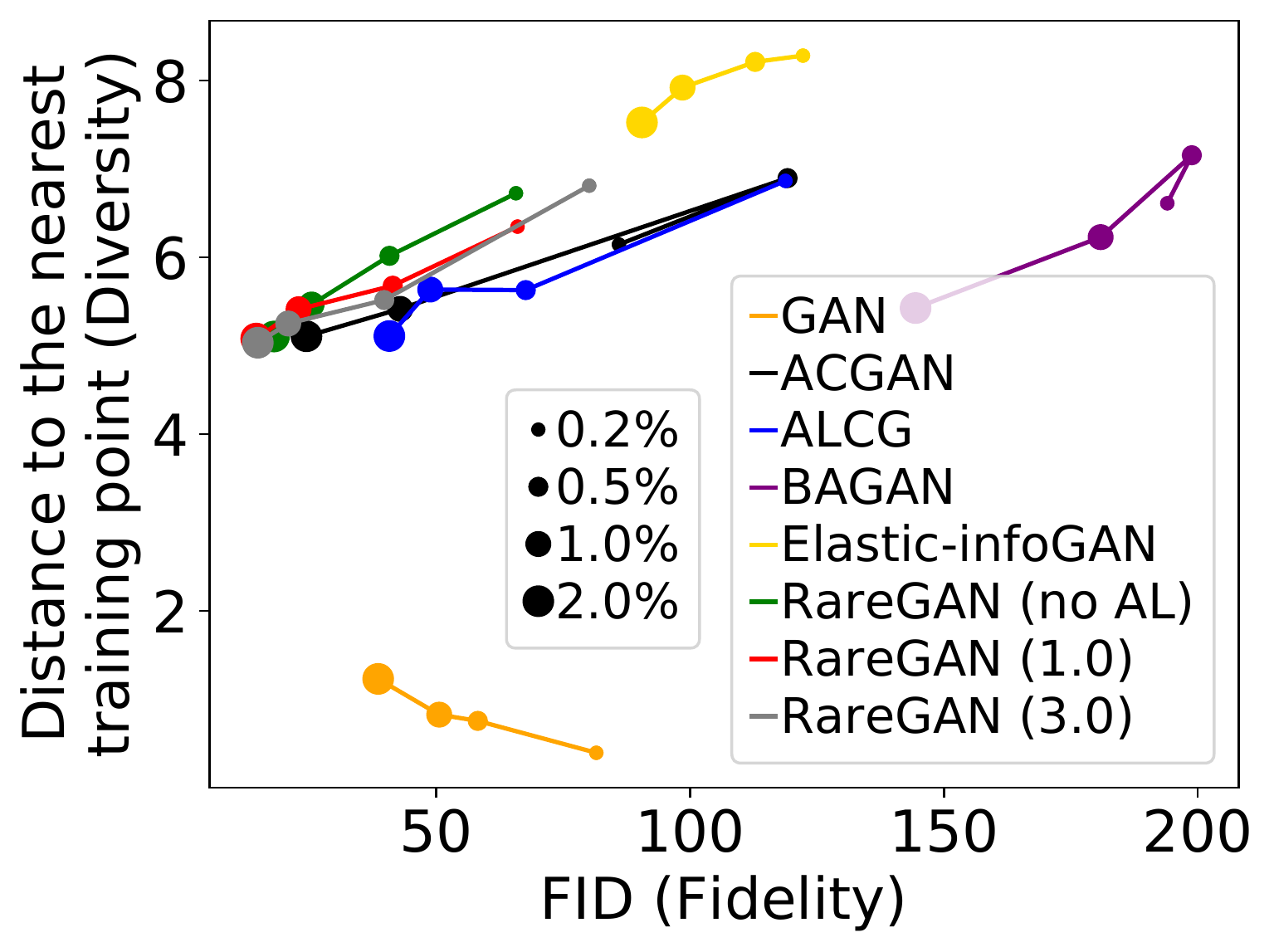}
			\vspace{-0.5cm}
			\caption{\mnist{} with varying rare class fraction. \name{} has the best  tradeoff.}
			\label{fig:low_frc_mnist}
		\end{minipage}
	\end{minipage}
\end{figure*}

\myparatightest{Robustness across applications}
We start with a qualitative comparison between baselines. 
We show randomly-generated samples on \mnist{} and \cifar{}.\footnote{All samples are drawn from the model with the median FID score over 5 runs.}
GAN produces high-quality \mnist{} images in \cref{fig:mnist_sample} by memorizing the labeled rare data (9 samples).
Other baselines do not memorize, but either 
produce mode-collapsed samples (e.g., BAGAN \cite{mariani2018bagan}), 
low-quality \emph{and} mode-collapsed samples (e.g., ACGAN \cite{odena2017conditional}, ALCG \cite{xie2019learning}),
or samples from wrong classes (e.g., Elastic-InfoGAN \cite{ojha2019elastic}).
\name{} produces samples that are of the same quality as GANs, but with better diversity. 
On  \cifar{}, \cref{fig:cifar_sample} shows for each baseline randomly-generated samples (top row) and the closest real samples (bottom row). 
Again, GAN memorizes the training data, ACGAN and ALCG have poor image quality, and \name{} trades off between the two (its sample quality is slightly worse than GAN, but its diversity is much better).

Quantitatively, \cref{fig:default} plots the  fidelity-diversity tradeoff of each baseline on our datasets. 
Lower fidelity scores (left) and higher diversity scores (upwards) are better.
The main takeaway of \cref{fig:default} is that \name{} has the best tradeoff in our experiments.
We discuss each method. \emph{(a) GAN.} In all cases, GANs have poor diversity due to memorization. %
In network applications, GAN fidelity is good due to overfitting. In the image datasets, FID is bad, as FID scores capture overfitting. 
\emph{(b) ALCG and BAGAN.} Generally, ALCG and BAGAN have much worse fidelity than other methods, consistent to the qualitative results.
\emph{(c) Elastic-InfoGAN.} Elastic-InfoGAN has higher diversity but much worse fidelity (\cref{fig:default_mnist}). Note that the higher diversity metric here is an artifact of Elastic-InfoGAN incorrectly generating digits that are mostly not 0 (\cref{fig:mnist_sample}), as it is not able to learn the boundary between rare and common classes well.
\emph{(d) ACGAN.} ACGAN has better diversity metrics and less overfitting than GAN, at the cost of sample fidelity. 
\emph{(e) Using unlabeled data.} 
Comparing ``\name{} (no AL)'' with ``ACGAN'', we see that unlabeled data significantly helps the image datasets, but not the network datasets. 
This may be because of problem dimensionality: the dimension of the images are much larger than the other two cases, so additional data gives a more prominent benefit. 
\emph{(f) Active learning and weighted loss.} 
Comparing \name{} (3.0 and 1.0) with \name{} (no AL), we see that weighted loss benefits  the network packet datasets, but not the  image datasets. This could be due to the complexity of the rare class boundary, which is  nonsmooth in network applications \cite{moon2021accurately}.

Due to space limitations, the following parametric studies show plots for a single dataset;
we defer the results on the other datasets to the appendices, where we see similar trends.

\myparatightest{Robustness to labeling budget $B$} 
We decrease $B$ to show how different approaches react in the low-budget setting. 
The results on \mnist{} are shown in \cref{fig:low_bgt_mnist}. 
All three \name{} versions are insensitive to budget. For the baselines, the sample qualities of ACGAN, ALCG, and BAGAN degrade significantly, as evidenced by the bad FIDs for small budgets. 
Elastic-InfoGAN has higher diversity again due to the incorrect generated digits.
GANs always have the worst diversity, no matter the budget. 
Results on other datasets are in \cref{app:bgt_frc}; \name{} generally has the best robustness across budgets.

\myparatightest{Robustness to rare class fraction $\rarefrc$} 
In \cref{fig:low_frc_mnist}, we vary $\rarefrc$ to measure the effect of class imbalance. 
All algorithms exhibit worse sample quality when the rare class fraction is decreased. 
However, for all $\rarefrc$, \name{} has a better fidelity-diversity tradeoff than ACGAN, ALCG, and BAGAN (achieving much better fidelity and similar diversity). 
Elastic-InfoGAN still has worse fidelity than \name{} due to wrong generated digits.
GANs always have the worst diversity.
Results on the other datasets are in \cref{app:bgt_frc}, where we see that \name{} generally has the best robustness to $\rarefrc$.

\myparatightest{Variance across trials} The standard error bars in  \cref{fig:default} show that ACGAN, ALCG, and BAGAN have high variance across trials, and \name{} with weighted loss has lower variance. 
This is because the weighted loss penalizes errors in the rare class, thus 
providing better stability.

The following ablation studies give additional insights into each tunable component of \name{}.

\myparatightest{Influence of the number of stages $\numstage$ (\cref{sec:approach_active}) and the loss weight $\weight$ (\cref{sec:approach_weight})} 
We have seen that active learning and weighted loss do not influence the image dataset results much (\cref{fig:default_mnist,fig:default_cifar,fig:low_bgt_mnist,fig:low_frc_mnist}). %
Therefore,  we focus on DNS in \cref{fig:weights_steps} (\cref{app:stages}). 
As we increase the weight from $w=1$ to $w=5$, both metrics improve, %
saturating at $w\geq 3$. 
At the default weight $w=3.0$, choosing $S=2$ or $S=4$ stages makes little difference.
Comparing \cref{fig:weights_steps} with \cref{fig:default_dns} \name{} improves upon ACGAN and ALCG for almost all $\numstage$ and $\weight$.

\myparatightest{Ablation study on \name{} components}
\name{} has three parts: (1) using unlabeled samples (\cref{sec:approach_unlabeled}), (2) active learning (\cref{sec:approach_active}), and (3) weighted loss (\cref{sec:approach_weight}). Active learning only makes sense with unlabeled samples, so there are 6 possible combinations. 
\cref{fig:comb} (\cref{app:ablation}) shows each variant on the DNS application. Including all components, \name{} yields the best diversity-fidelity tradeoff and low variance.

\myparatightest{Comparison to domain-specific techniques}
We compare to AmpMAP, the state-of-the-art work on (DNS) amplification attacks in the security community \cite{moon2021accurately}, in \cref{tbl:ampmap}.  AmpMAP finds high amplification packets by drawing random packets and requesting their amplification factors, and then doing random field perturbation on high amplification packets. AmpMAP uses amplification threshold 10, and the same packet space as ours.  
Note that AmpMAP is speciﬁcally designed for ampliﬁcation attacks, not applicable for other applications we did. Even in that case, our proposed \name{} still 
achieves much better fidelity and diversity with a fraction of the budget.

%% file: tex/discussion.tex
\section{Conclusions}
\label{sec:conclusion}

We propose \name{} for generating samples from a rare class subject to a limited labeling budget.
We show that \name{} has good, stable diversity and fidelity in experiments covering different loss functions (e.g., Jensen-Shannon divergence \cite{goodfellow2014generative}, Wasserstein distance \cite{arjovsky2017wasserstein,gulrajani2017improved}), architectures (e.g., CNN, MLP), data types (e.g., network packets, images),  budgets, and rare class fractions.

%% file: tex/ack.tex
\section*{Acknowledgements}
We thank Yucheng Yin for the help on baseline comparison, and Sekar Kulandaivel, Wenyu Wang, Bryan Phee, and Shruti Datta Gupta for their help with earlier versions of \name{}.
This work was supported in part by faculty research awards from Google, JP Morgan Chase, and the Sloan Foundation, as well as gift grants from Cisco and Siemens AG.
This research was sponsored in part by National Science Foundation Convergence Accelerator award 2040675 and the U.S. Army Combat
Capabilities Development Command Army Research Laboratory and was accomplished under Cooperative Agreement
Number W911NF-13-2-0045 (ARL Cyber Security CRA).
The views and conclusions contained in this document are
those of the authors and should not be interpreted as representing the official policies, either expressed or implied, of the
Combat Capabilities Development Command Army Research
Laboratory or the U.S. Government. The U.S. Government
is authorized to reproduce and distribute reprints for Government purposes notwithstanding any copyright notation here
on.
Zinan Lin acknowledges the support of the Siemens FutureMakers Fellowship, the CMU Presidential Fellowship, and the Cylab Presidential Fellowship.
This work used the Extreme Science and Engineering Discovery Environment (XSEDE) \cite{xsede}, which is supported by National Science Foundation grant number ACI-1548562. Specifically, it used the Bridges system \cite{bridges}, which is supported by NSF award number ACI-1445606, at the Pittsburgh Supercomputing Center (PSC).

%% file: tex/app_proof.tex
\section{Proof of \cref{thm:unlabeled}}
\label{app:proof_unlabeled}

Let $\p^*_c$ be $\p^*$ under condition ``common''. Note that the objective function in \cref{thm:unlabeled} is the ACGAN loss, which has two parts: the GAN loss and the classifier loss; recall the ACGAN classifier loss in \cref{eq:classification_loss}.

First, suppose that  $d(\p^*_c,\pcommon)=0$ and $d(\p^*_r,\prare)=0$. 
It is clear that in this case, $p^*$ optimizes the GAN loss $d(\p^*,p)$. 
Since $\mathcal C$ contains all measurable functions and since $\support{\prare} \cap\support{\pcommon}=\emptyset$, 
there exists a classifier $C\in \mathcal C$ that can separate the rare class $\prare$ from the common one $\pcommon$. 
Since $\p^*_r=\prare$ and $\p^*_c=\pcommon$, the same $C$ also has 0 loss on $p^*$, so the classifier loss is 0.
Hence we have that $\p^*$ optimizes both parts of the AGCAN loss. 

Now suppose that either $d(\p^*_c,\pcommon)>0$ or $d(\p^*_r,\prare)>0$. Then we have the following possibilities:

(Case 1) $\support{\p_c^*} - \support{\p_c}\not=\emptyset$ or $\support{\p_r^*} - \support{\p_r}\not=\emptyset$. 
In this case, we know the classification loss cannot achieve the optimal value %
(namely 0).
Since either $\p_c^*$ contains elements not in $\p_c$ or $\p_r^*$ contains elements not in $\p_r$, any classifier $C$ must either have nonzero error on the rare or the common class.
Hence, the ACGAN classifier loss will be nonzero. 

(Case 2) $\support{\p_c^*} \subseteq \support{\p_c}$ and $\support{\p_r^*} \subseteq \support{\p_r}$. 
In this case, because $d(\p^*_c,\pcommon)>0$ and/or $d(\p^*_r,\prare)>0$,
the GAN loss cannot achieve its optimal value of 0. 

Therefore, we can conclude that for any $\p^* \in \arg\min_{\pgenerate} \min_C d\bra{\pgenerate,\p} + \lclassification\bra{C,\pgeneratesamplelabel;\psamplelabel'}$, we have  $d\bra{\p_r^*,\p_r}=0$.

\myparatightest{Comments} The key here is the assumption $\support{\prare} \cap\support{\pcommon}=\emptyset$ (stated in problem statement \cref{sec:intro}). If this is not satisfied, the bias in the labeled dataset (because of active learning) will influence the final solution through the classification loss. To be more specific, even if $\p^*$ satisfies $d(\p^*_c,\pcommon)=0$ and $d(\p^*_r,\prare)=0$, it does not necessarily optimize the classification loss. Therefore, the optimal solution could be moved away from this $\p^*$.

\section{Proof of \cref{thm:weight}}
\label{app:proof_weight}
It suffices to show that 
\begin{align}
 \Eb_{x\sim \p} \brb{W(x)\cdot \log D(x)} &= \Eb_{x\sim \q} \brb{ \log D(x)}\label{eq:prove1}\\
 \frac{1}{s} \cdot\Eb_{x\sim \pgenerate} \brb{W(x)\cdot\log \bra{1-D(x)}}&=\Eb_{x\sim \qgenerate} \brb{\log \bra{1-D(x)}}\label{eq:prove2}\\
 \Eb_{x\sim \p} \brb{ W(x)\cdot D(x)} &= \Eb_{x\sim \q} \brb{ D(x)}\label{eq:prove3}\\
  \frac{1}{s}\cdot \Eb_{x\sim \pgenerate} \brb{ W(x) \cdot D(x)}&=\Eb_{x\sim \qgenerate} \brb{ D(x)}\label{eq:prove4}
\end{align}

We start from \cref{eq:prove3}. 
\begin{align*}
	 &\quad\Eb_{x\sim \p} \brb{ W(x)\cdot D(x)} \\
	&=\int \p(x) W(x) D(x)dx\\
	&=\int_{\support{\prare}}\rarefrc\weight \cdot \prare(x) D(x)dx + \int_{\support{\pcommon}} (1-\rarefrc) \frac{1-\weight\rarefrc}{1-\rarefrc} \cdot \pcommon\bra{x}D(x)dx\\
	&=\int_{\support{\prare}}\rarefrc\weight \cdot \prare(x) D(x)dx + \int_{\support{\pcommon}} (1-\weight \rarefrc) \cdot \pcommon\bra{x}D(x)dx\\
	&=\quad\Eb_{x\sim \q} \brb{ W(x)\cdot D(x)}
\end{align*}
The proof for \cref{eq:prove1} is similar except changing $D(x)$ to $\log D(x)$.

Now we look at \cref{eq:prove4}.
\begin{align*}
	&\quad \frac{1}{s}\cdot \Eb_{x\sim \pgenerate} \brb{ W(x) \cdot D(x)}\\
&=\frac{1}{s}\bra{ \int \pgenerate(x)W(x)D(x)dx }\\
&=\frac{1}{s}\bra{ 
	\int_{\support{\prare}} \rarefrclearned \weight\cdot \pgeneraterare(x)D(x)dx +
\int_{X-\support{\prare}} (1-\rarefrclearned)  \frac{1-\weight\rarefrc}{1-\rarefrc} \cdot \pgeneratecommon(x) D(x)dx}\\
&=\Eb_{x\sim \qgenerate} \brb{ D(x)}
\end{align*}
The proof for \cref{eq:prove2} is similar except changing $D(x)$ to $\log \bra{1-D(x)}$.

%% file: tex/experiment_details.tex
\section{Experiment details}
\label{app:experiment_details}
\subsection{DNS amplification attacks}
The details of the packet fields are as follows (following \cite{moon2021accurately,lin2019towards}).
\begin{itemize}
	\item  id: 16 bits, modeled using 16 2-dim softmax
	\item opcode: choosing from [0,1,2,3,4,5], modeled using a 5-dim softmax
	\item aa: 1 bit, modeled using one 2-dim softmax
	\item tc: 1 bit, modeled using one 2-dim softmax
	\item rd: 1 bit, modeled using one 2-dim softmax
	\item ra: 1 bit, modeled using one 2-dim softmax
	\item z: 1 bit, modeled using one 2-dim softmax
	\item ad: 1 bit, modeled using one 2-dim softmax
	\item cd: 1 bit, modeled using one 2-dim softmax
	\item rcode: 4 bits, modeled using 4 2-dim softmax
	\item rdatatype: choosing from [1, 28, 18, 42, 257, 60, 59, 37, 5, 49, 32769, 39, 48, 43, 55, 45, 25, 36, 29, 15, 35, 2, 47, 50, 51, 61, 12, 46, 17, 24, 6, 33, 44, 32768, 249, 52, 250, 16, 256, 255, 252, 251, 41], modeled using one 43-dim softmax
	\item Rdataclass: choosing from [1,3,4,255], modeled using one 4-dim softmax
	\item edns: 1 bit, modeled using one 2-dim softmax
	\item dnssec: 1 bit, modeled using one 2-dim softmax
	\item payload: 16 bits, modeled using 16 2-dim softmax
	\item url: choosing from ['berkeley.edu', 'energy.gov', 'chase.com', 'aetna.com', 'google.com', 'Nairaland.com', 'Alibaba.com', 'Cambridge.org', 'Alarabiya.net', 'Bnamericas.com'], modeled using one 10-dim softmax
\end{itemize}

\subsection{Packet classification}

The details of the packet fields are as follows.
\begin{itemize}
	\item  Source IP: 32 bits, modeled using 32 2-dim softmax
	\item  Destination IP: 32 bits, modeled using 32 2-dim softmax
	\item  Source port: 16 bits, modeled using 16 2-dim softmax
	\item  Destination port: 16 bits, modeled using 16 2-dim softmax
	\item  Protocol: 7 bits, modeled using 7 2-dim softmax
\end{itemize}

\subsection{BAGAN evaluation}
To make a fair comparison with BAGAN, we have taken the following approach, which is similar to how we compared with GAN/ACGAN in the paper. First, we request the labels of randomly (uniformly) selected samples up to the given budget ($B$), and then use those labeled samples to train BAGAN. We used the official BAGAN for MNIST GitHub repo \url{https://github.com/IBM/BAGAN}, and the only modifications we made are: (1) instead of having 10 classes, we treat the digit 0 as the rare class, and all other digits as the common class (consistent to how we compared other baselines in paper), and (2) we trained on the labeled samples described above. We keep all other hyper-parameters the same as the original code. 

\subsection{Elastic-InfoGAN evaluation}
We take the official Elastic-infoGAN for MNIST code from \url{https://github.com/utkarshojha/elastic-infogan}, and do the following modifications to make it suitable for our problem:
\begin{itemize}
	\item InfoGAN loss: We keep the InfoGAN loss (which uses all unlabeled data), and on top of it, we use the labeled data to train Q (InfoGAN encoder) in a supervised way.
	\item Contrastive loss: For a randomly selected labeled sample, we randomly pick a sample with the same label as the positive sample, and a sample with the different label as the negative sample, and apply the same contrastive loss (treating the cosine similarity divided by a temperature parameter as logits) on top of the positive/negative pairs.
	\item Instead of having 10 classes, we treat the digit 0 as the rare class, and all other digits as the common class (consistent to how we compared other baselines in paper).
\end{itemize} 
We keep all other hyper-parameters the same as the original code.

\subsection{Computation resources}
All the experiments were run on a public cluster: Bridges-2 system at the Pittsburgh Supercomputing Center (PSC) with NVIDIA Tesla V100 GPUs. All the experiments took around 10k GPU hours. To evaluate DNS amplifications, we set up our own DNS resolvers on Cloudlab \cite{duplyakin2019design}. The server parameters are:
\begin{verbatim}
	CPU: Eight 64-bit ARMv8 (Atlas/A57) cores at 2.4 GHz (APM X-GENE)
	RAM: 64GB ECC Memory (8x 8 GB DDR3-1600 SO-DIMMs)
	Disk: 120 GB of flash (SATA3 / M.2, Micron M500)
	NIC: Dual-port Mellanox ConnectX-3 10 GB NIC (PCIe v3.0, 8 lanes)
\end{verbatim}
To evaluate the packet classification time, we set up our own evaluation nodes on Cloudlab \cite{duplyakin2019design}. The server parameters are:
\begin{verbatim}
	CPU: 64-bit Intel Quad Core Xeon E5530
	RAM: 12GB
\end{verbatim}

%% file: tex/app_bgt_frc.tex
\section{Additional Results: Robustness w.r.t. Labeling Budget and Rare Class Fraction}
\label{app:bgt_frc}

\begin{figure}[h]
	\centering
	\includegraphics[width=0.35\linewidth]{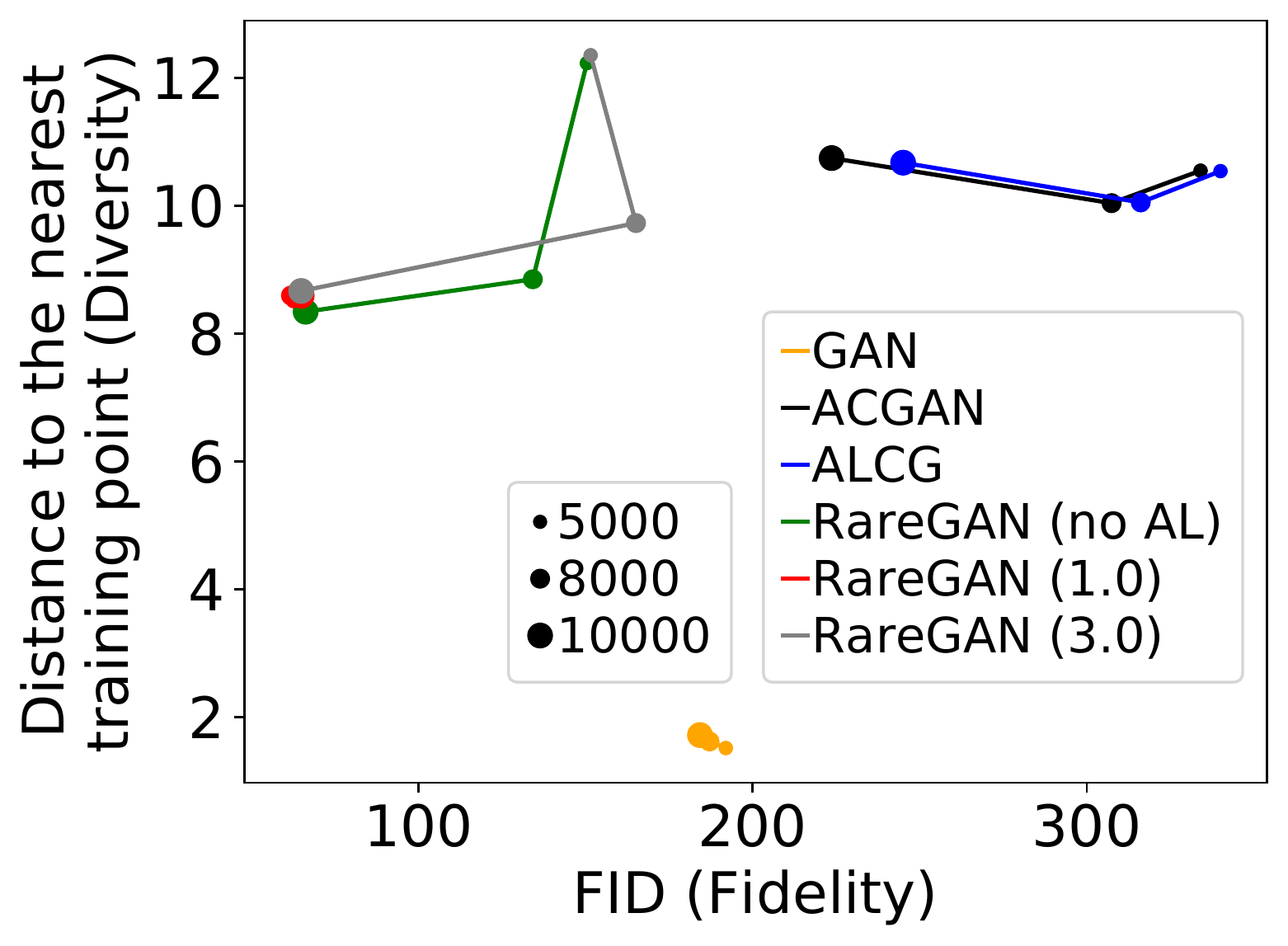}
	\caption{\cifar{} with different budgets.}
	\label{fig:low_bgt_cifar}
\end{figure}

\begin{figure}[h]
	\centering
	\includegraphics[width=0.35\linewidth]{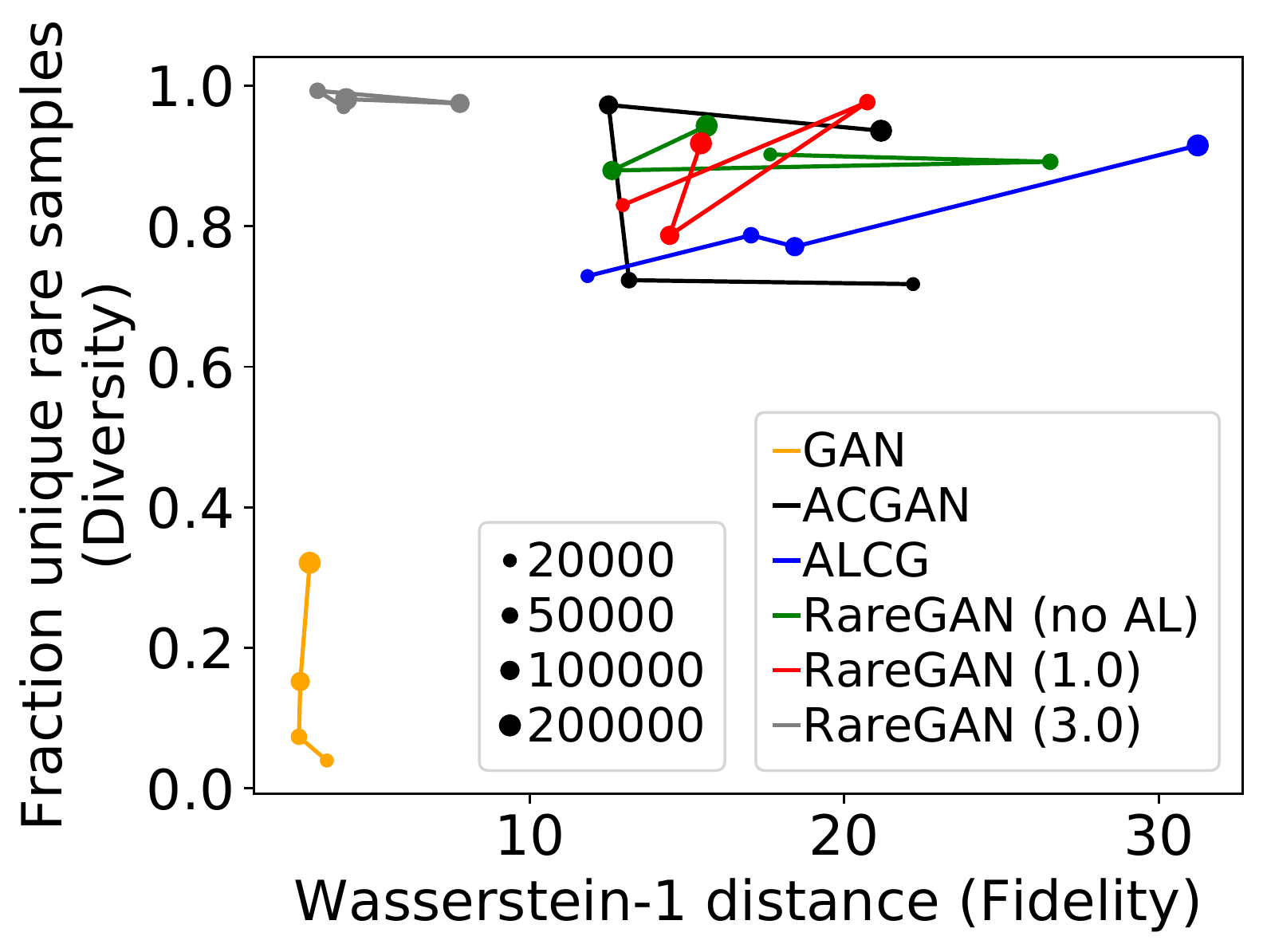}
	\caption{DNS amplification attacks with different budgets.}
	\label{fig:low_bgt_dns}
\end{figure}

\begin{figure}[h]
	\centering
	\includegraphics[width=0.35\linewidth]{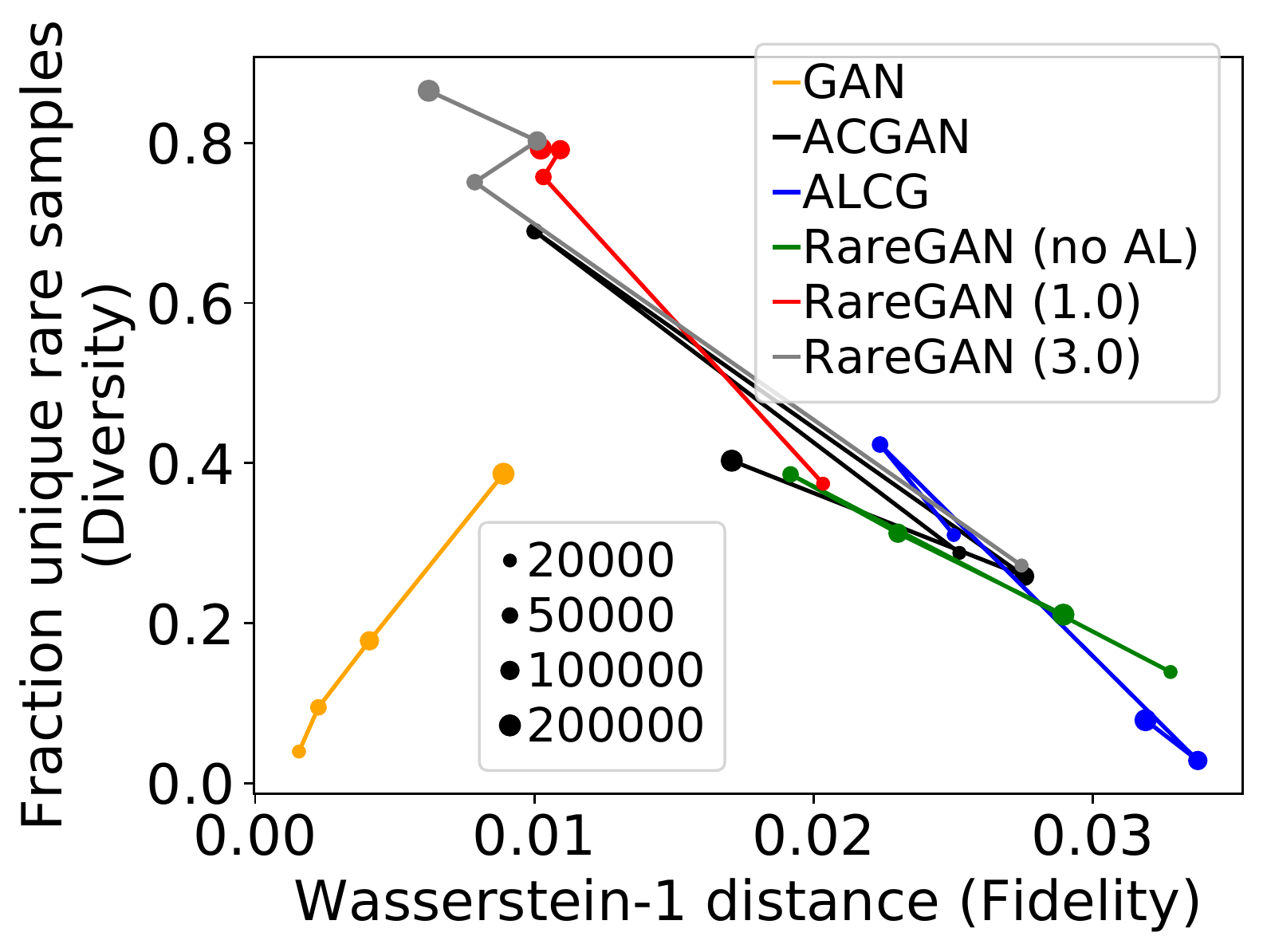}
	\caption{Packet classifiers with different budgets.}
	\label{fig:low_bgt_pc}
\end{figure}

\begin{figure}[h]
	\centering
	\includegraphics[width=0.35\linewidth]{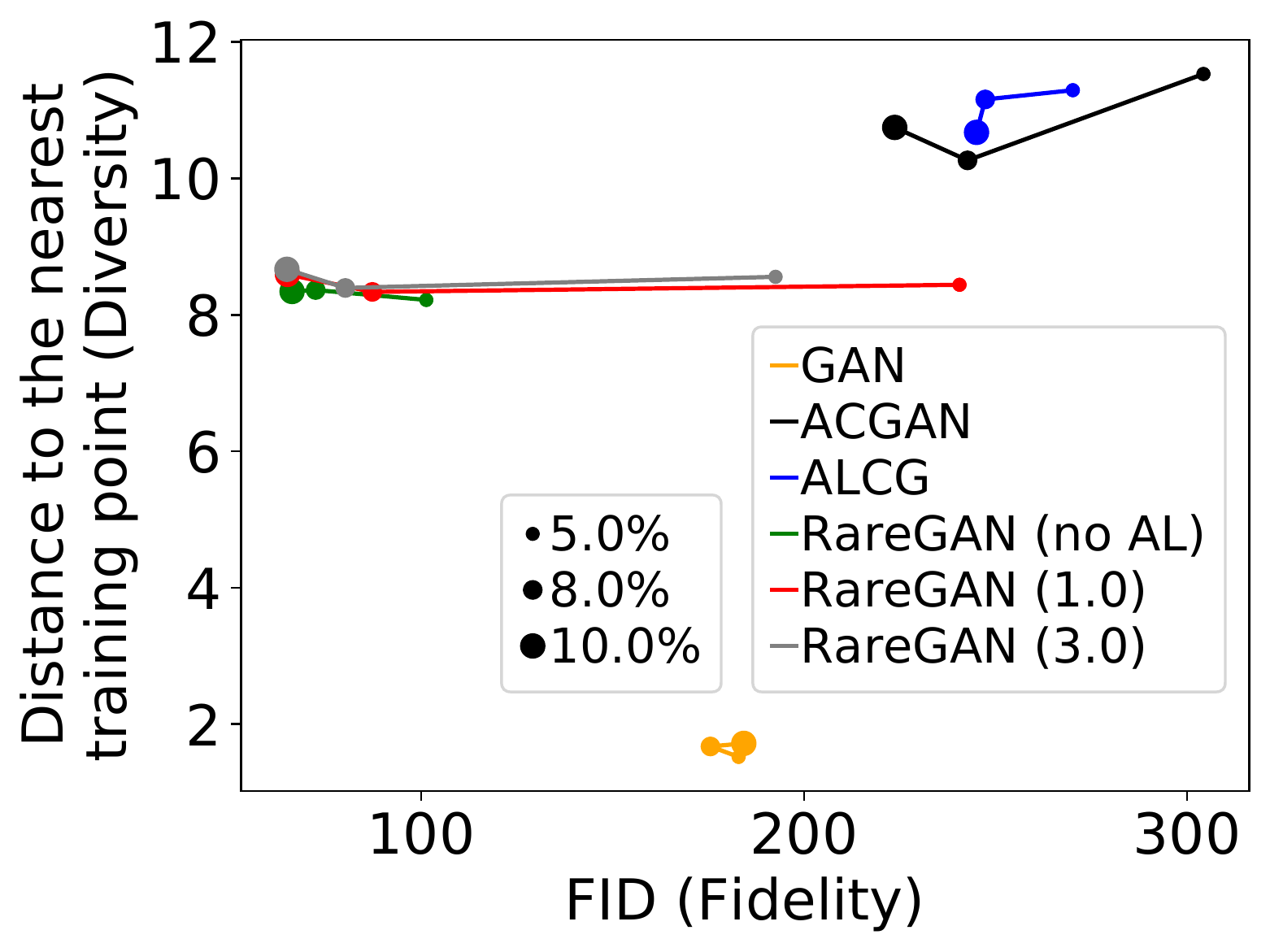}
	\caption{\cifar{} with different rare class fractions.}
	\label{fig:low_frc_cifar}
\end{figure}

\begin{figure}[h]
	\centering
	\includegraphics[width=0.35\linewidth]{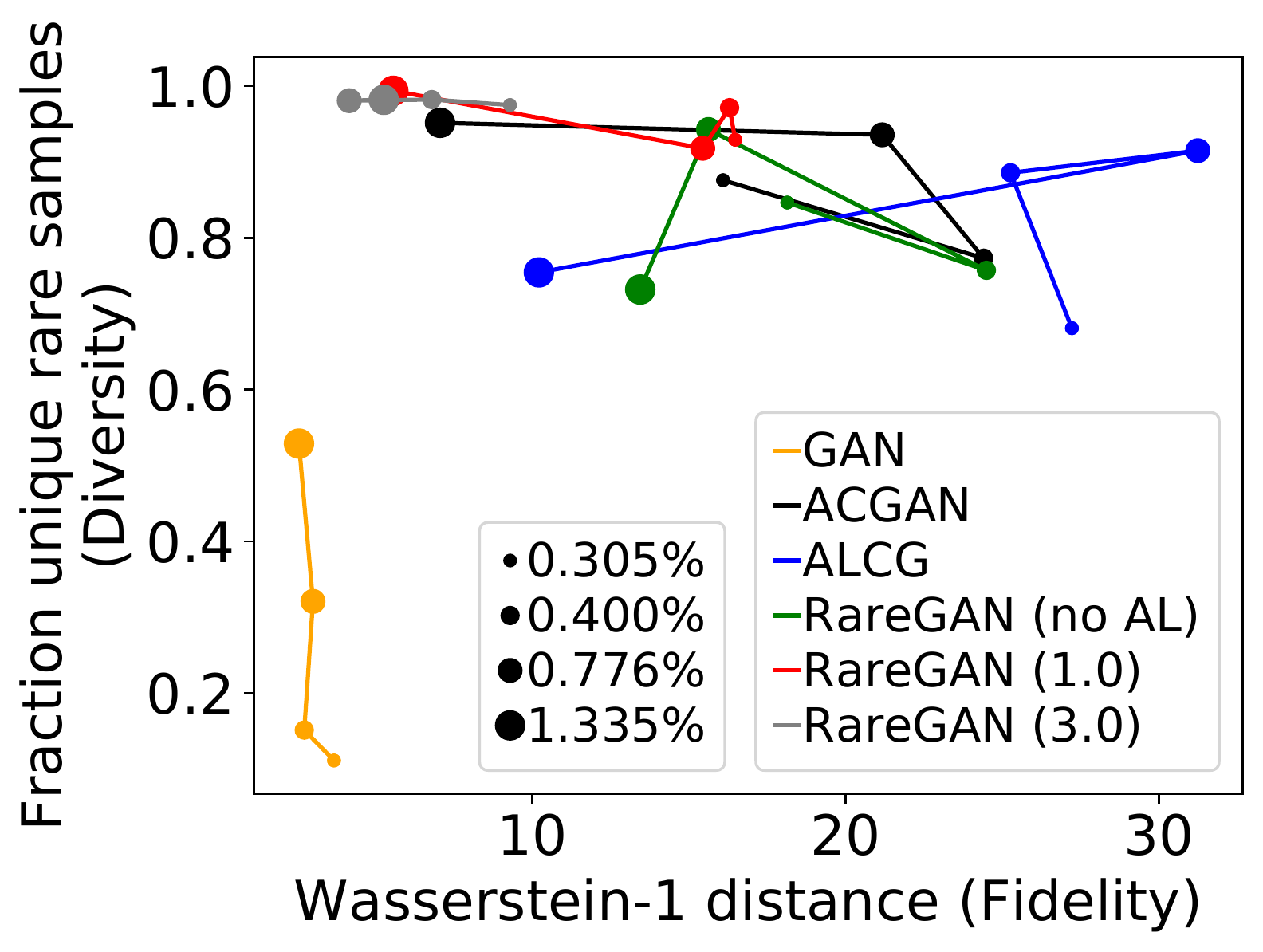}
	\caption{DNS amplification attacks with different rare class fractions. The fractions correspond to $T=20,15,10,5$ respectively.}
	\label{fig:low_frc_dns}
\end{figure}

\begin{figure}[h]
	\centering
	\includegraphics[width=0.35\linewidth]{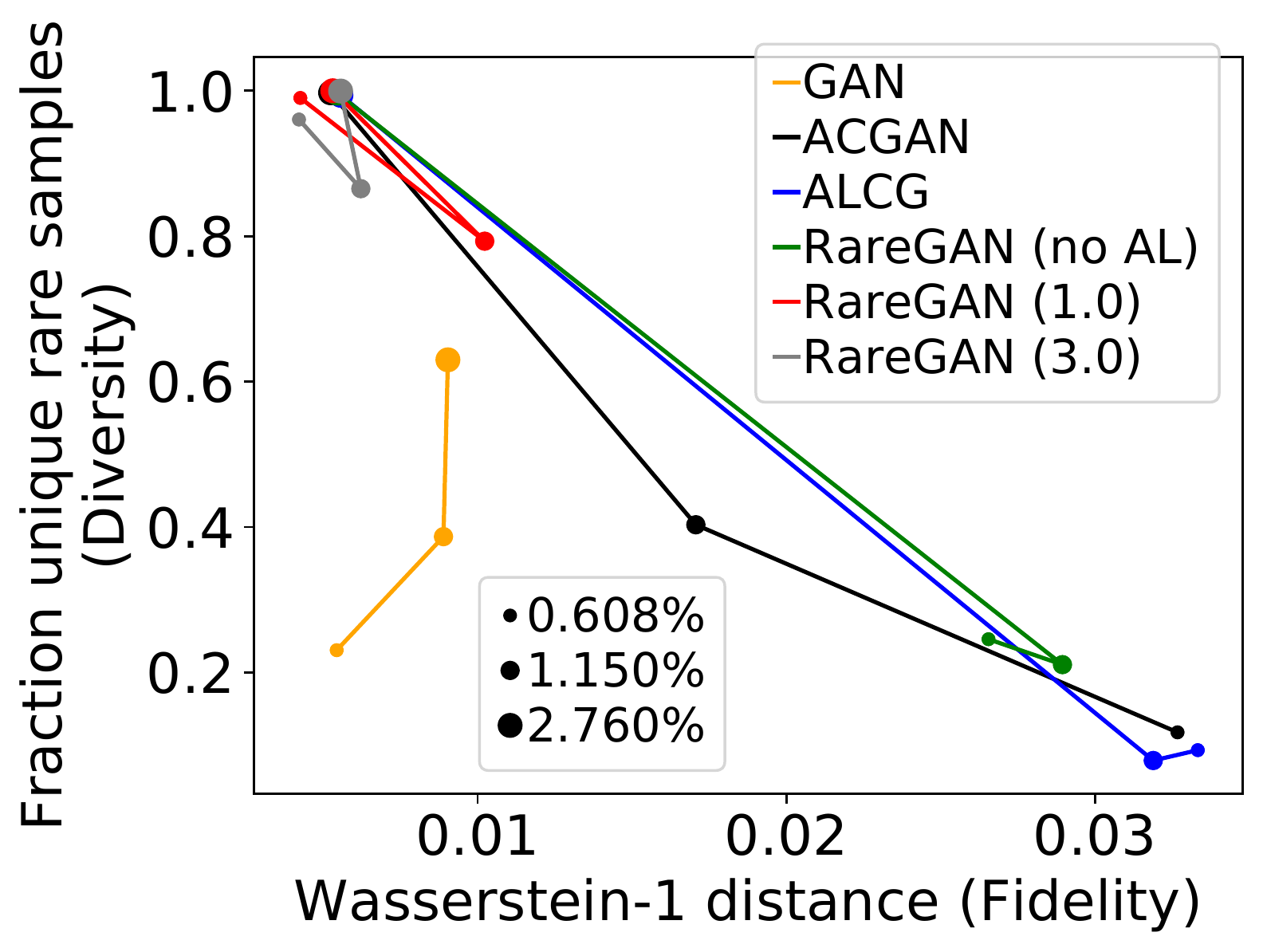}
	\caption{Packet classifiers with different rare class fractions. The fractions correspond to $T=0.06,0.055,0.05$ respectively.}
	\label{fig:low_frc_pc}
\end{figure}

\begin{figure}[h]
	\centering
	\includegraphics[width=0.35\linewidth]{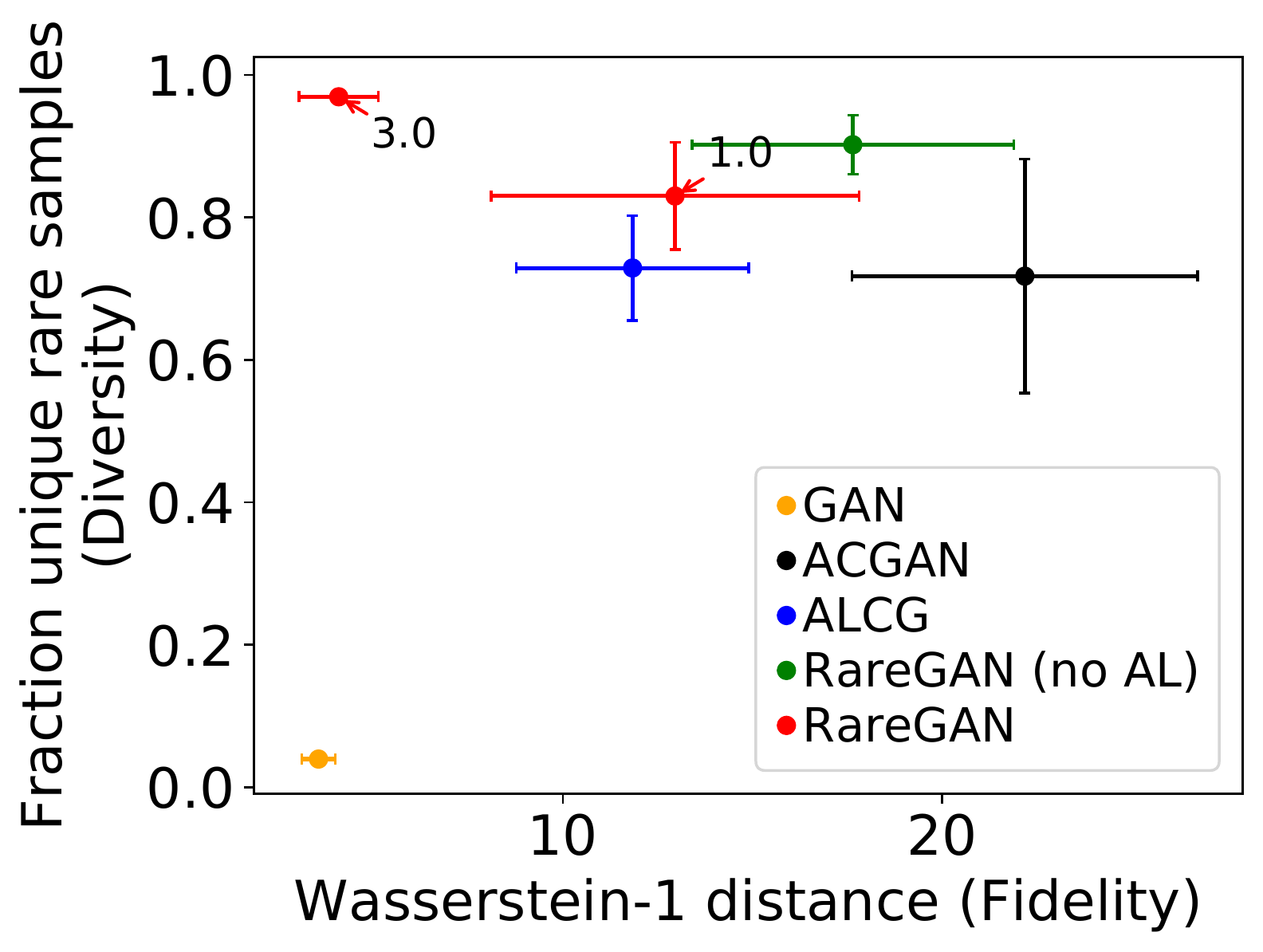}
	\caption{DNS amplification attacks with $B=\numprint{20000}$.}
	\label{fig:low_bgt_dns_20000}
\end{figure}

\begin{figure}
	\centering
	\includegraphics[width=0.35\linewidth]{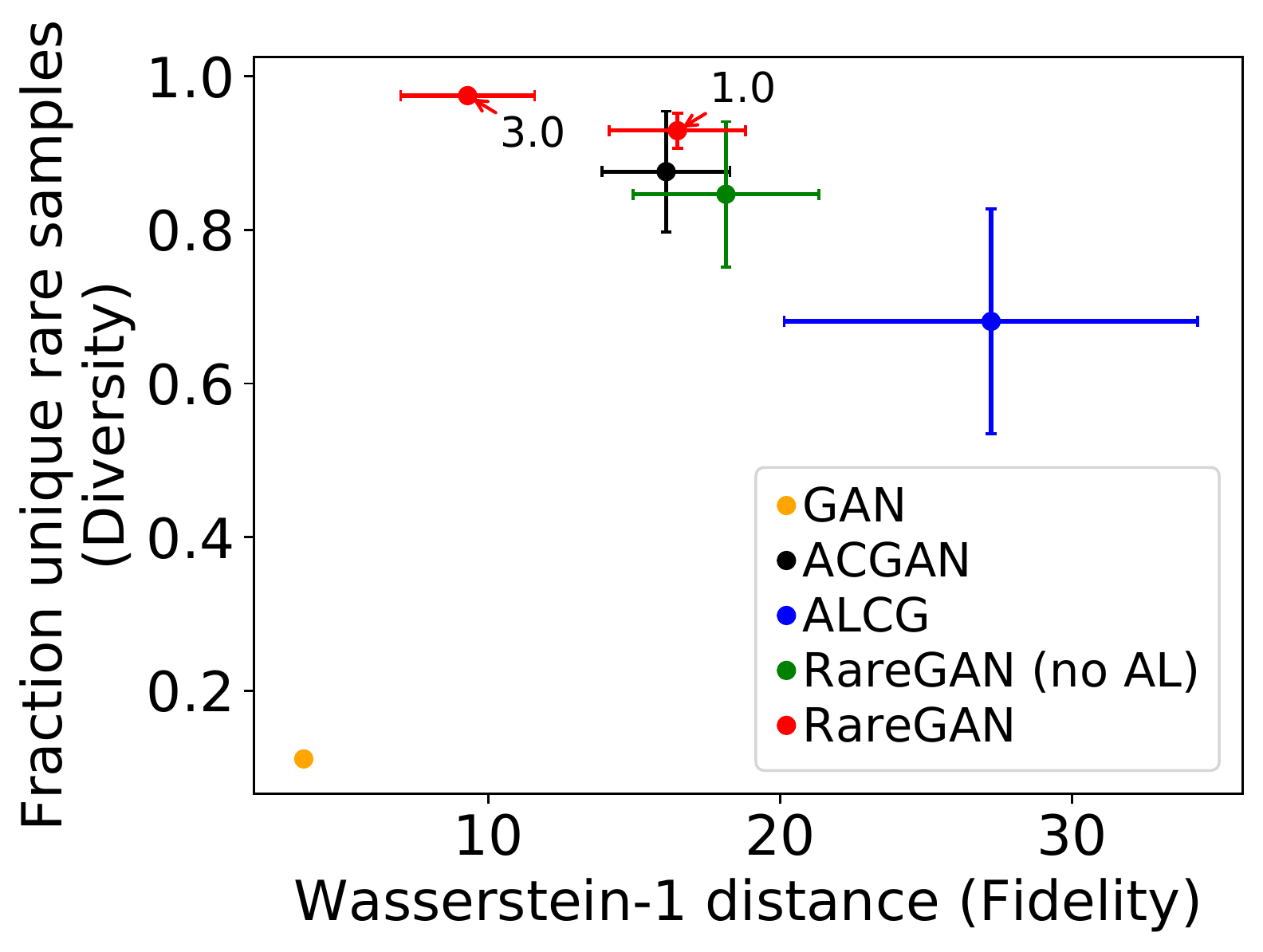}
	\caption{DNS amplication attacks with $\alpha=0.305\%$ (corresponding to $T=20$).}
	\label{fig:low_frc_dns_20}
\end{figure}

The results with different labeling budget on \mnist{}, \cifar{}, DNS amplification attacks, and packet classifiers are in \cref{fig:low_bgt_mnist,fig:low_bgt_cifar,fig:low_bgt_dns,fig:low_bgt_pc} respectively. 
The results with different rare class fractions are on \mnist{}, \cifar{}, DNS amplification attacks, and packet classifiers are in \cref{fig:low_frc_mnist,fig:low_frc_cifar,fig:low_frc_dns,fig:low_frc_pc} respectively.
In all cases, we can see that \name{} with active learning and weighted loss improves upon the baselines. 
Two other observations: (1) We see that the curves in \cref{fig:low_bgt_dns,fig:low_frc_dns} do not have a clear pattern. The reason is that the random trials in DNS tend to have large variances. To illustrate this, we pick the smallest budget in \cref{fig:low_bgt_dns} and the smallest rare class fraction in \cref{fig:low_frc_dns} and plot their error bars in \cref{fig:low_bgt_dns_20000,fig:low_frc_dns_20}. We can see that ACGAN, ALCG, and \name{} without weighted loss all have large variances. However, weighted loss is able to reduce the variance a lot (as we already demonstrated in the main text).
(2) We can see that in \cifar{} all algorithms tend to have bad FIDs when the budget or rare class fraction are small, as this dataset is challenging. However, even in that case, \name{} versions still have better FIDs than the baselines.


\section{Additional Results: Effect of stages $S$ and loss weight $w$}
\label{app:stages}
\begin{figure}[h]
	\centering
	\includegraphics[width=0.45\linewidth]{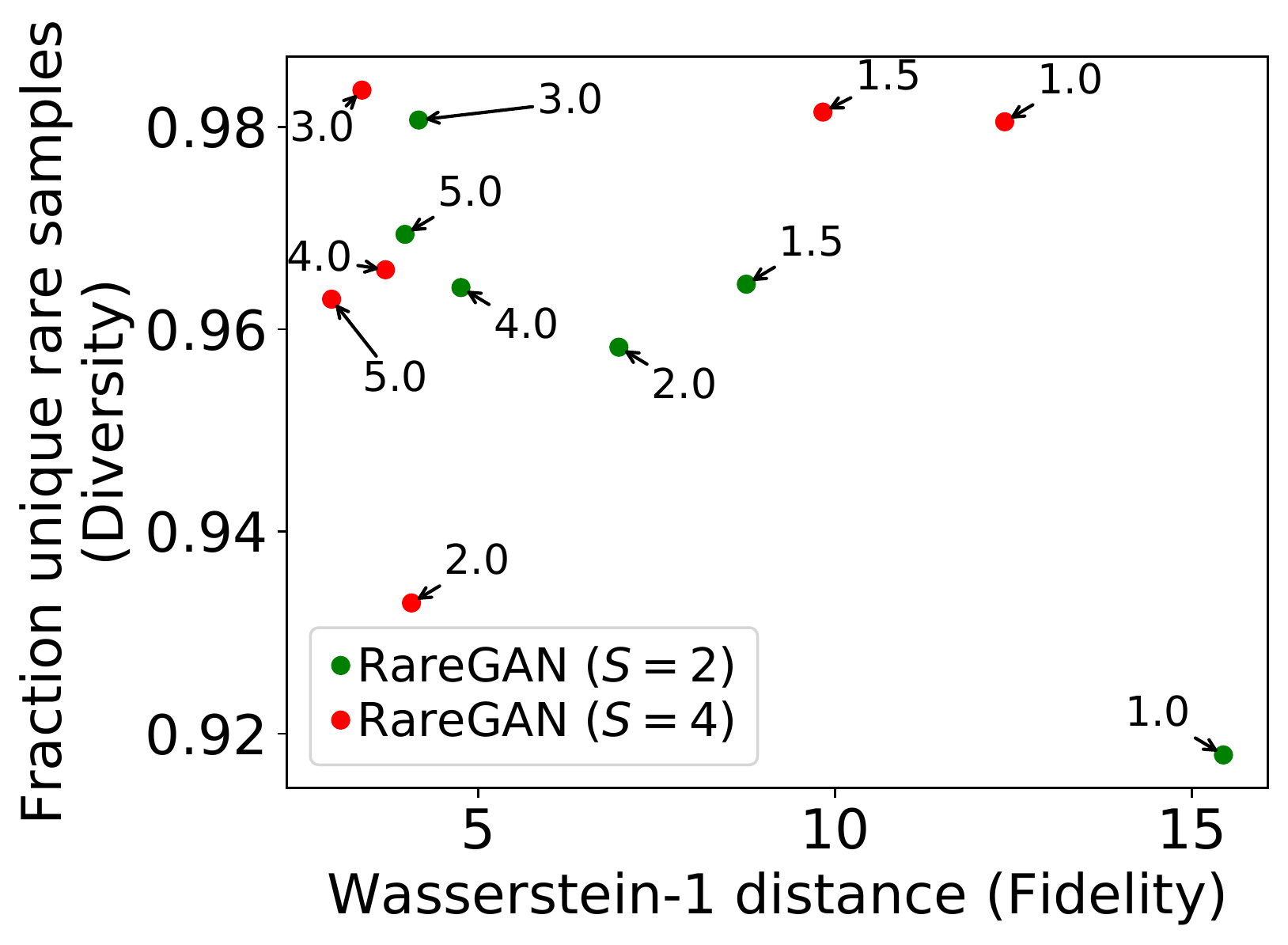}
	\caption{\name{} with different weights $\weight$ and numbers of stages $\numstage$ on DNS amplification attacks.}
	\label{fig:weights_steps}
\end{figure}

Figure \ref{fig:weights_steps} illustrates the effect of the two main hyperparameters of \name: the number of stages $S$ and the weight $w$.

\section{Additional Results: Ablation Study}
\label{app:ablation}
\begin{figure}[h]
	\centering
	\includegraphics[width=0.45\linewidth]{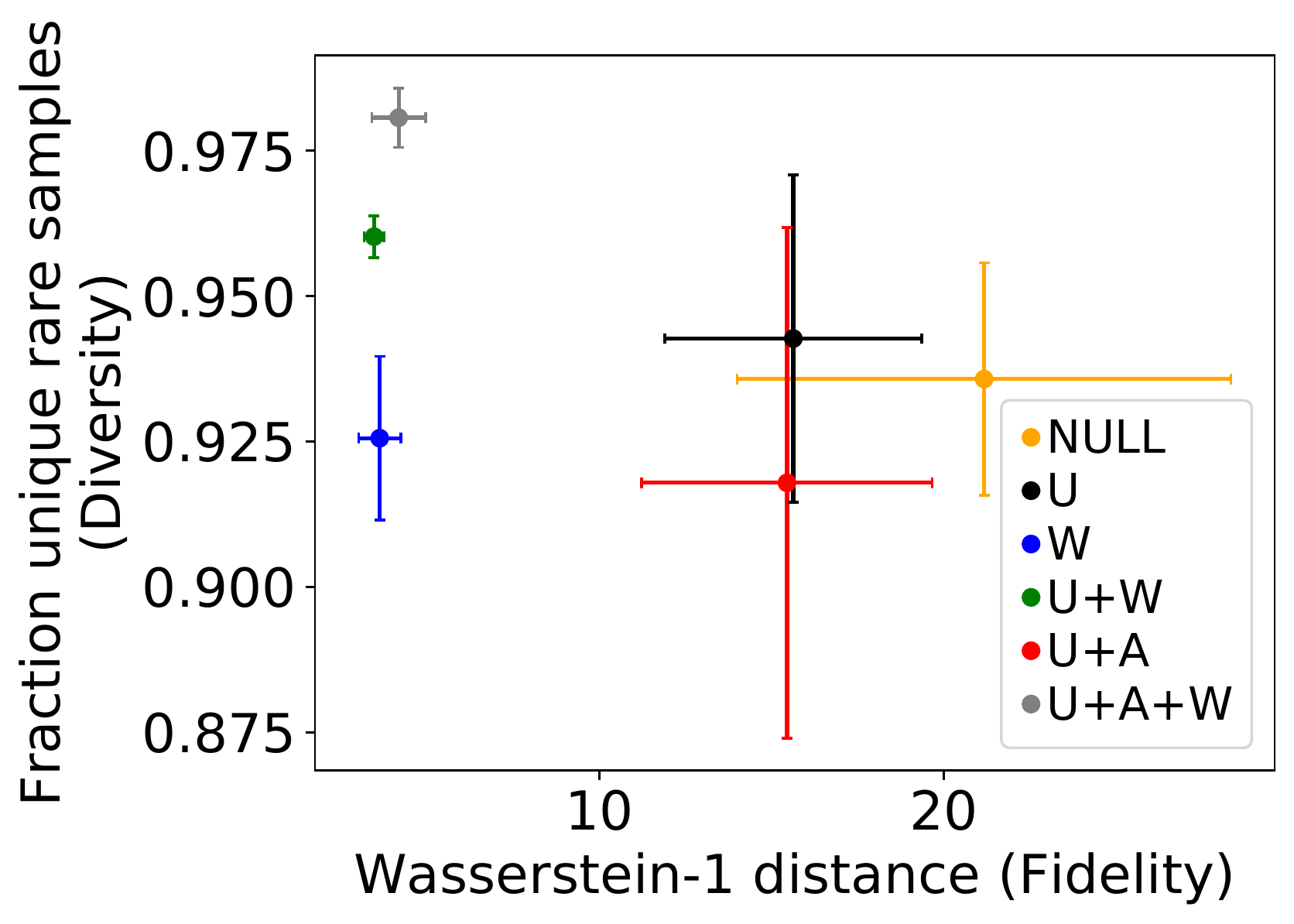}
	\caption{Different combinations of \name{} components on DNS amplification attacks. U, A, W, NULL refer to using unlabeled samples, active learning, weighted loss, and none of them respectively. Bars show standard error over 5 runs.}
	\label{fig:comb}
	\vspace{-0.5cm}
\end{figure}

Figure \ref{fig:comb} illustrates the effect of using various sub-components of \name on the DNS dataset. We observe that the full \name{} has the best utility-diversity tradeoff.

%% file: main.bbl
\begin{thebibliography}{66}
\providecommand{\natexlab}[1]{#1}

\bibitem[{Ali-Gombe and Elyan(2019)}]{ali2019mfc}
Ali-Gombe, A.; and Elyan, E. 2019.
\newblock MFC-GAN: class-imbalanced dataset classification using multiple fake
  class generative adversarial network.
\newblock \emph{Neurocomputing}, 361.

\bibitem[{Anagnostopoulos et~al.(2013)Anagnostopoulos, Kambourakis, Kopanos,
  Louloudakis, and Gritzalis}]{anagnostopoulos2013dns}
Anagnostopoulos, M.; Kambourakis, G.; Kopanos, P.; Louloudakis, G.; and
  Gritzalis, S. 2013.
\newblock DNS amplification attack revisited.
\newblock \emph{Computers \& Security}, 39: 475--485.

\bibitem[{Arjovsky, Chintala, and Bottou(2017)}]{arjovsky2017wasserstein}
Arjovsky, M.; Chintala, S.; and Bottou, L. 2017.
\newblock Wasserstein generative adversarial networks.
\newblock In \emph{ICML}, 214--223. PMLR.

\bibitem[{Arora and Zhang(2017)}]{arora2017gans}
Arora, S.; and Zhang, Y. 2017.
\newblock Do gans actually learn the distribution? an empirical study.
\newblock \emph{arXiv preprint arXiv:1706.08224}.

\bibitem[{Asokan and Seelamantula(2020)}]{asokan2020teaching}
Asokan, S.; and Seelamantula, C.~S. 2020.
\newblock Teaching a gan what not to learn.
\newblock \emph{arXiv preprint arXiv:2010.15639}.

\bibitem[{Augenstein et~al.(2019)Augenstein, McMahan, Ramage, Ramaswamy,
  Kairouz, Chen, Mathews et~al.}]{augenstein2019generative}
Augenstein, S.; McMahan, H.~B.; Ramage, D.; Ramaswamy, S.; Kairouz, P.; Chen,
  M.; Mathews, R.; et~al. 2019.
\newblock Generative models for effective ML on private, decentralized
  datasets.
\newblock \emph{arXiv preprint arXiv:1911.06679}.

\bibitem[{Black, Guttman, and Okun(2021)}]{black2021guidelines}
Black, P.~E.; Guttman, B.; and Okun, V. 2021.
\newblock Guidelines on Minimum Standards for Developer Verification of
  Software.
\newblock arXiv:2107.12850.

\bibitem[{Caballero et~al.(2007)Caballero, Yin, Liang, and
  Song}]{caballero2007polyglot}
Caballero, J.; Yin, H.; Liang, Z.; and Song, D. 2007.
\newblock Polyglot: Automatic extraction of protocol message format using
  dynamic binary analysis.
\newblock In \emph{Proceedings of the 14th ACM conference on Computer and
  communications security}.

\bibitem[{Chen et~al.(2019)Chen, Zhai, Ritter, Lucic, and
  Houlsby}]{chen2019self}
Chen, T.; Zhai, X.; Ritter, M.; Lucic, M.; and Houlsby, N. 2019.
\newblock Self-supervised gans via auxiliary rotation loss.
\newblock In \emph{CVPR}, 12154--12163.

\bibitem[{Chen et~al.(2016)Chen, Duan, Houthooft, Schulman, Sutskever, and
  Abbeel}]{chen2016infogan}
Chen, X.; Duan, Y.; Houthooft, R.; Schulman, J.; Sutskever, I.; and Abbeel, P.
  2016.
\newblock Infogan: Interpretable representation learning by information
  maximizing generative adversarial nets.
\newblock In \emph{Proceedings of the 30th International Conference on Neural
  Information Processing Systems}, 2180--2188.

\bibitem[{Chiu et~al.(2018)Chiu, Ruan, Shen, and Hung}]{chiu2018design}
Chiu, Y.-K.; Ruan, S.-J.; Shen, C.-A.; and Hung, C.-C. 2018.
\newblock The design and implementation of a latency-aware packet
  classification for OpenFlow protocol based on FPGA.
\newblock In \emph{Proceedings of the 2018 VII International Conference on
  Network, Communication and Computing}, 64--69.

\bibitem[{Choi et~al.(2018)Choi, Choi, Kim, Ha, Kim, and
  Choo}]{choi2018stargan}
Choi, Y.; Choi, M.; Kim, M.; Ha, J.-W.; Kim, S.; and Choo, J. 2018.
\newblock Stargan: Unified generative adversarial networks for multi-domain
  image-to-image translation.
\newblock In \emph{CVPR}, 8789--8797.

\bibitem[{Cui et~al.(2019)Cui, Jia, Lin, Song, and Belongie}]{cui2019class}
Cui, Y.; Jia, M.; Lin, T.-Y.; Song, Y.; and Belongie, S. 2019.
\newblock Class-balanced loss based on effective number of samples.
\newblock In \emph{CVPR}, 9268--9277.

\bibitem[{Dai et~al.(2017)Dai, Yang, Yang, Cohen, and
  Salakhutdinov}]{dai2017good}
Dai, Z.; Yang, Z.; Yang, F.; Cohen, W.~W.; and Salakhutdinov, R. 2017.
\newblock Good semi-supervised learning that requires a bad gan.
\newblock \emph{arXiv preprint arXiv:1705.09783}.

\bibitem[{Douzas and Bacao(2018)}]{douzas2018effective}
Douzas, G.; and Bacao, F. 2018.
\newblock Effective data generation for imbalanced learning using conditional
  generative adversarial networks.
\newblock \emph{Expert Systems with applications}, 91: 464--471.

\bibitem[{Duplyakin et~al.(2019)Duplyakin, Ricci, Maricq, Wong, Duerig, Eide,
  Stoller, Hibler, Johnson, Webb et~al.}]{duplyakin2019design}
Duplyakin, D.; Ricci, R.; Maricq, A.; Wong, G.; Duerig, J.; Eide, E.; Stoller,
  L.; Hibler, M.; Johnson, D.; Webb, K.; et~al. 2019.
\newblock The design and operation of CloudLab.
\newblock In \emph{2019 {USENIX} Annual Technical Conference}, 1--14.

\bibitem[{Goodfellow et~al.(2014)Goodfellow, Pouget-Abadie, Mirza, Xu,
  Warde-Farley, Ozair, Courville, and Bengio}]{goodfellow2014generative}
Goodfellow, I.~J.; Pouget-Abadie, J.; Mirza, M.; Xu, B.; Warde-Farley, D.;
  Ozair, S.; Courville, A.; and Bengio, Y. 2014.
\newblock Generative adversarial networks.
\newblock \emph{arXiv preprint arXiv:1406.2661}.

\bibitem[{Gulrajani et~al.(2017)Gulrajani, Ahmed, Arjovsky, Dumoulin, and
  Courville}]{gulrajani2017improved}
Gulrajani, I.; Ahmed, F.; Arjovsky, M.; Dumoulin, V.; and Courville, A. 2017.
\newblock Improved training of stein gans.
\newblock \emph{arXiv preprint arXiv:1704.00028}.

\bibitem[{Haque(2020)}]{haque2020ec}
Haque, A. 2020.
\newblock EC-GAN: Low-Sample Classification using Semi-Supervised Algorithms
  and GANs.
\newblock \emph{arXiv preprint arXiv:2012.15864}.

\bibitem[{Heusel et~al.(2017)Heusel, Ramsauer, Unterthiner, Nessler, and
  Hochreiter}]{heusel2017gans}
Heusel, M.; Ramsauer, H.; Unterthiner, T.; Nessler, B.; and Hochreiter, S.
  2017.
\newblock Gans trained by a two time-scale update rule converge to a local nash
  equilibrium.
\newblock \emph{arXiv preprint arXiv:1706.08500}.

\bibitem[{hwalsuklee(2018)}]{acgangithub}
hwalsuklee. 2018.
\newblock tensorflow-generative-model-collections.
\newblock
  \url{https://github.com/hwalsuklee/tensorflow-generative-model-collections}.

\bibitem[{Joshi, Porikli, and Papanikolopoulos(2009)}]{joshi2009multi}
Joshi, A.~J.; Porikli, F.; and Papanikolopoulos, N. 2009.
\newblock Multi-class active learning for image classification.
\newblock In \emph{CVPR}, 2372--2379. IEEE.

\bibitem[{Kambourakis et~al.(2007)Kambourakis, Moschos, Geneiatakis, and
  Gritzalis}]{kambourakis2007fair}
Kambourakis, G.; Moschos, T.; Geneiatakis, D.; and Gritzalis, S. 2007.
\newblock A fair solution to dns amplification attacks.
\newblock In \emph{Second International Workshop on Digital Forensics and
  Incident Analysis (WDFIA 2007)}, 38--47. IEEE.

\bibitem[{Karras et~al.(2020)Karras, Aittala, Hellsten, Laine, Lehtinen, and
  Aila}]{karras2020training}
Karras, T.; Aittala, M.; Hellsten, J.; Laine, S.; Lehtinen, J.; and Aila, T.
  2020.
\newblock Training generative adversarial networks with limited data.
\newblock \emph{arXiv preprint arXiv:2006.06676}.

\bibitem[{Kong et~al.(2019)Kong, Tong, Klinkigt, Watanabe, Akira, and
  Murakami}]{kong2019active}
Kong, Q.; Tong, B.; Klinkigt, M.; Watanabe, Y.; Akira, N.; and Murakami, T.
  2019.
\newblock Active generative adversarial network for image classification.
\newblock In \emph{AAAI}, volume~33, 4090--4097.

\bibitem[{Krizhevsky(2009)}]{cifar}
Krizhevsky, A. 2009.
\newblock Learning multiple layers of features from tiny images.

\bibitem[{Kumar, Sattigeri, and Fletcher(2017)}]{kumar2017semi}
Kumar, A.; Sattigeri, P.; and Fletcher, T. 2017.
\newblock Semi-supervised learning with gans: Manifold invariance with improved
  inference.
\newblock \emph{Advances in Neural Information Processing Systems}, 30.

\bibitem[{LeCun et~al.(1998)LeCun, Bottou, Bengio, and Haffner}]{mnist}
LeCun, Y.; Bottou, L.; Bengio, Y.; and Haffner, P. 1998.
\newblock Gradient-based learning applied to document recognition.
\newblock \emph{Proceedings of the IEEE}, 86(11): 2278--2324.

\bibitem[{Lewis and Catlett(1994)}]{lewis1994heterogeneous}
Lewis, D.~D.; and Catlett, J. 1994.
\newblock Heterogeneous uncertainty sampling for supervised learning.
\newblock In \emph{Machine learning proceedings 1994}, 148--156. Elsevier.

\bibitem[{Li and Sethi(2006)}]{li2006confidence}
Li, M.; and Sethi, I.~K. 2006.
\newblock Confidence-based active learning.
\newblock \emph{IEEE transactions on pattern analysis and machine
  intelligence}, 28(8): 1251--1261.

\bibitem[{Liang et~al.(2019)Liang, Zhu, Jin, and Stoica}]{liang2019neural}
Liang, E.; Zhu, H.; Jin, X.; and Stoica, I. 2019.
\newblock Neural packet classification.
\newblock In \emph{Proceedings of the ACM Special Interest Group on Data
  Communication}, 256--269.

\bibitem[{Liang et~al.(2020)Liang, Yu, Xu, Raj, and
  Singh}]{liang2020controlled}
Liang, H.; Yu, L.; Xu, G.; Raj, B.; and Singh, R. 2020.
\newblock Controlled AutoEncoders to Generate Faces from Voices.
\newblock In \emph{International Symposium on Visual Computing}, 476--487.
  Springer.

\bibitem[{Lin et~al.(2019)Lin, Moon, Zarate, Mulagalapalli, Kulandaivel, Fanti,
  and Sekar}]{lin2019towards}
Lin, Z.; Moon, S.-J.; Zarate, C.~M.; Mulagalapalli, R.; Kulandaivel, S.; Fanti,
  G.; and Sekar, V. 2019.
\newblock Towards oblivious network analysis using generative adversarial
  networks.
\newblock In \emph{Proceedings of the 18th ACM Workshop on Hot Topics in
  Networks}, 43--51.

\bibitem[{Lin et~al.(2020)Lin, Thekumparampil, Fanti, and Oh}]{lin2020infogan}
Lin, Z.; Thekumparampil, K.; Fanti, G.; and Oh, S. 2020.
\newblock Infogan-cr and modelcentrality: Self-supervised model training and
  selection for disentangling gans.
\newblock In \emph{International Conference on Machine Learning}, 6127--6139.
  PMLR.

\bibitem[{Mao et~al.(2017)Mao, Li, Xie, Lau, Wang, and
  Paul~Smolley}]{mao2017least}
Mao, X.; Li, Q.; Xie, H.; Lau, R.~Y.; Wang, Z.; and Paul~Smolley, S. 2017.
\newblock Least squares generative adversarial networks.
\newblock In \emph{ICCV}, 2794--2802.

\bibitem[{Mariani et~al.(2018)Mariani, Scheidegger, Istrate, Bekas, and
  Malossi}]{mariani2018bagan}
Mariani, G.; Scheidegger, F.; Istrate, R.; Bekas, C.; and Malossi, C. 2018.
\newblock Bagan: Data augmentation with balancing gan.
\newblock \emph{arXiv preprint arXiv:1803.09655}.

\bibitem[{Matwyshyn et~al.(2010)Matwyshyn, Cui, Keromytis, and
  Stolfo}]{ethicssecurity}
Matwyshyn, A.~M.; Cui, A.; Keromytis, A.~D.; and Stolfo, S.~J. 2010.
\newblock { Ethics in Security Vulnerability Research}.
\newblock In \emph{IEEE Security \& Privacy}.

\bibitem[{Mirza and Osindero(2014)}]{mirza2014conditional}
Mirza, M.; and Osindero, S. 2014.
\newblock Conditional generative adversarial nets.
\newblock \emph{arXiv preprint arXiv:1411.1784}.

\bibitem[{Moon et~al.(2021)Moon, Yin, Sharma, Yuan, Spring, and
  Sekar}]{moon2021accurately}
Moon, S.-J.; Yin, Y.; Sharma, R.~A.; Yuan, Y.; Spring, J.~M.; and Sekar, V.
  2021.
\newblock Accurately measuring global risk of amplification attacks using
  ampmap.
\newblock In \emph{30th {USENIX} Security Symposium}.

\bibitem[{Mullick, Datta, and Das(2019)}]{mullick2019generative}
Mullick, S.~S.; Datta, S.; and Das, S. 2019.
\newblock Generative adversarial minority oversampling.
\newblock In \emph{ICCV}, 1695--1704.

\bibitem[{Naeem et~al.(2020)Naeem, Oh, Uh, Choi, and Yoo}]{naeem2020reliable}
Naeem, M.~F.; Oh, S.~J.; Uh, Y.; Choi, Y.; and Yoo, J. 2020.
\newblock Reliable fidelity and diversity metrics for generative models.
\newblock In \emph{ICML}, 7176--7185. PMLR.

\bibitem[{Nowozin, Cseke, and Tomioka(2016)}]{nowozin2016f}
Nowozin, S.; Cseke, B.; and Tomioka, R. 2016.
\newblock f-gan: Training generative neural samplers using variational
  divergence minimization.
\newblock In \emph{NeurIPS}.

\bibitem[{Nystrom et~al.(2015)Nystrom, Levine, Roskies, and Scott}]{bridges}
Nystrom, N.~A.; Levine, M.~J.; Roskies, R.~Z.; and Scott, J.~R. 2015.
\newblock Bridges: A Uniquely Flexible HPC Resource for New Communities and
  Data Analytics.
\newblock In \emph{Proceedings of the 2015 XSEDE Conference: Scientific
  Advancements Enabled by Enhanced Cyberinfrastructure}, XSEDE '15, 30:1--30:8.
  New York, NY, USA: ACM.
\newblock ISBN 978-1-4503-3720-5.

\bibitem[{Odena(2016)}]{odena2016semi}
Odena, A. 2016.
\newblock Semi-supervised learning with generative adversarial networks.
\newblock \emph{arXiv preprint arXiv:1606.01583}.

\bibitem[{Odena, Olah, and Shlens(2017)}]{odena2017conditional}
Odena, A.; Olah, C.; and Shlens, J. 2017.
\newblock Conditional image synthesis with auxiliary classifier gans.
\newblock In \emph{ICML}. PMLR.

\bibitem[{Ojha et~al.(2019)Ojha, Singh, Hsieh, and Lee}]{ojha2019elastic}
Ojha, U.; Singh, K.~K.; Hsieh, C.-J.; and Lee, Y.~J. 2019.
\newblock Elastic-InfoGAN: Unsupervised Disentangled Representation Learning in
  Class-Imbalanced Data.
\newblock \emph{arXiv preprint arXiv:1910.01112}.

\bibitem[{Pedrosa et~al.(2018)Pedrosa, Iyer, Zaostrovnykh, Fietz, and
  Argyraki}]{castan}
Pedrosa, L.; Iyer, R.; Zaostrovnykh, A.; Fietz, J.; and Argyraki, K. 2018.
\newblock Automated synthesis of adversarial workloads for network functions.
\newblock In \emph{Proceedings of the 2018 Conference of the ACM Special
  Interest Group on Data Communication}.

\bibitem[{Petsios et~al.(2017)Petsios, Zhao, Keromytis, and
  Jana}]{petsios2017slowfuzz}
Petsios, T.; Zhao, J.; Keromytis, A.~D.; and Jana, S. 2017.
\newblock Slowfuzz: Automated domain-independent detection of algorithmic
  complexity vulnerabilities.
\newblock In \emph{Proceedings of the 2017 ACM SIGSAC Conference on Computer
  and Communications Security}, 2155--2168.

\bibitem[{Rangwani, Mopuri, and Radhakrishnan(2021)}]{rangwani2021class}
Rangwani, H.; Mopuri, K.~R.; and Radhakrishnan, V.~B. 2021.
\newblock Class Balancing {GAN} with a Classifier in the Loop.

\bibitem[{Rashelbach, Rottenstreich, and
  Silberstein(2020)}]{rashelbach2020computational}
Rashelbach, A.; Rottenstreich, O.; and Silberstein, M. 2020.
\newblock A Computational Approach to Packet Classification.
\newblock In \emph{Proceedings of the Annual conference of the ACM Special
  Interest Group on Data Communication on the applications, technologies,
  architectures, and protocols for computer communication}, 542--556.

\bibitem[{Ren, Liu, and Liu(2019)}]{ren2019ewgan}
Ren, J.; Liu, Y.; and Liu, J. 2019.
\newblock EWGAN: Entropy-based Wasserstein GAN for imbalanced learning.
\newblock In \emph{AAAI}, volume~33, 10011--10012.

\bibitem[{Rossow(2014)}]{rossow2014amplification}
Rossow, C. 2014.
\newblock Amplification Hell: Revisiting Network Protocols for DDoS Abuse.
\newblock In \emph{NDSS}.

\bibitem[{Salimans et~al.(2016)Salimans, Goodfellow, Zaremba, Cheung, Radford,
  and Chen}]{salimans2016improved}
Salimans, T.; Goodfellow, I.; Zaremba, W.; Cheung, V.; Radford, A.; and Chen,
  X. 2016.
\newblock Improved techniques for training gans.
\newblock \emph{arXiv preprint arXiv:1606.03498}.

\bibitem[{Shmelkov, Schmid, and Alahari(2018)}]{shmelkov2018good}
Shmelkov, K.; Schmid, C.; and Alahari, K. 2018.
\newblock How good is my GAN?
\newblock In \emph{ECCV}, 213--229.

\bibitem[{Sivaraman and Trivedi(2010)}]{sivaraman2010general}
Sivaraman, S.; and Trivedi, M.~M. 2010.
\newblock A general active-learning framework for on-road vehicle recognition
  and tracking.
\newblock \emph{IEEE Transactions on Intelligent Transportation Systems},
  11(2): 267--276.

\bibitem[{Soylu, Erdem, and Carus(2020)}]{soylu2020bit}
Soylu, T.; Erdem, O.; and Carus, A. 2020.
\newblock Bit vector-coded simple CART structure for low latency traffic
  classification on FPGAs.
\newblock \emph{Computer Networks}, 167: 106977.

\bibitem[{Sun, Bhattarai, and Kim(2020)}]{sun2020matchgan}
Sun, J.; Bhattarai, B.; and Kim, T.-K. 2020.
\newblock MatchGAN: a self-supervised semi-supervised conditional generative
  adversarial network.
\newblock In \emph{Proceedings of the Asian Conference on Computer Vision}.

\bibitem[{{Towns} et~al.(2014){Towns}, {Cockerill}, {Dahan}, {Foster},
  {Gaither}, {Grimshaw}, {Hazlewood}, {Lathrop}, {Lifka}, {Peterson},
  {Roskies}, {Scott}, and {Wilkins-Diehr}}]{xsede}
{Towns}, J.; {Cockerill}, T.; {Dahan}, M.; {Foster}, I.; {Gaither}, K.;
  {Grimshaw}, A.; {Hazlewood}, V.; {Lathrop}, S.; {Lifka}, D.; {Peterson},
  G.~D.; {Roskies}, R.; {Scott}, J.~R.; and {Wilkins-Diehr}, N. 2014.
\newblock XSEDE: Accelerating Scientific Discovery.
\newblock \emph{Computing in Science Engineering}, 16(5): 62--74.

\bibitem[{Vlachos(2008)}]{vlachos2008stopping}
Vlachos, A. 2008.
\newblock A stopping criterion for active learning.
\newblock \emph{Computer Speech \& Language}, 22(3): 295--312.

\bibitem[{Wang, Zhang, and Van De~Weijer(2016)}]{wang2016ensembles}
Wang, Y.; Zhang, L.; and Van De~Weijer, J. 2016.
\newblock Ensembles of generative adversarial networks.
\newblock \emph{arXiv preprint arXiv:1612.00991}.

\bibitem[{Wei et~al.(2019)Wei, Suriawinata, Vaickus, Ren, Liu, Wei, and
  Hassanpour}]{wei2019generative}
Wei, J.; Suriawinata, A.; Vaickus, L.; Ren, B.; Liu, X.; Wei, J.; and
  Hassanpour, S. 2019.
\newblock Generative image translation for data augmentation in colorectal
  histopathology images.
\newblock \emph{arXiv preprint arXiv:1910.05827}.

\bibitem[{Xie and Huang(2019)}]{xie2019learning}
Xie, M.-K.; and Huang, S.-J. 2019.
\newblock Learning class-conditional gans with active sampling.
\newblock In \emph{Proceedings of the 25th ACM SIGKDD International Conference
  on Knowledge Discovery \& Data Mining}, 998--1006.

\bibitem[{Yang and Zhou(2021)}]{yang2021ida}
Yang, H.; and Zhou, Y. 2021.
\newblock IDA-GAN: A Novel Imbalanced Data Augmentation GAN.
\newblock In \emph{ICPR}. IEEE.

\bibitem[{Zadorozhnyy, Cheng, and Ye(2021)}]{zadorozhnyy2021adaptive}
Zadorozhnyy, V.; Cheng, Q.; and Ye, Q. 2021.
\newblock Adaptive Weighted Discriminator for Training Generative Adversarial
  Networks.
\newblock In \emph{Proceedings of the IEEE/CVF Conference on Computer Vision
  and Pattern Recognition}, 4781--4790.

\bibitem[{Zhao et~al.(2020)Zhao, Liu, Lin, Zhu, and
  Han}]{zhao2020differentiable}
Zhao, S.; Liu, Z.; Lin, J.; Zhu, J.-Y.; and Han, S. 2020.
\newblock Differentiable augmentation for data-efficient gan training.
\newblock \emph{arXiv preprint arXiv:2006.10738}.

\bibitem[{Zhou et~al.(2018)Zhou, Liu, Zhou, and Chen}]{zhou2018gan}
Zhou, T.; Liu, W.; Zhou, C.; and Chen, L. 2018.
\newblock Gan-based semi-supervised for imbalanced data classification.
\newblock In \emph{ICIM}. IEEE.

\end{thebibliography}
